\documentclass[preprint,12pt]{elsarticle}
\usepackage{amssymb}
\usepackage{amsmath}
\usepackage{lineno}

\usepackage{hyperref}
\hypersetup{
	colorlinks=true,
	linkcolor=blue,
	filecolor=magenta,
	urlcolor=cyan,
}
\usepackage{amsmath}
\usepackage{amssymb}
\usepackage{mathrsfs}
\usepackage{amsthm}
\usepackage{cleveref}
\newtheorem{thm}{Theorem}[section]
\newtheorem{Thm}[thm]{Theorem}

\theoremstyle{definition}

\theoremstyle{remark}
\newtheorem*{Rem}{Remark}

\crefname{Thm}{Theorem}{Theorems}
\crefname{Rem}{Remark}{Remarks}
\usepackage{bm}
\usepackage{enumerate}

\usepackage{algorithm}
\usepackage{algpseudocode}
\usepackage{booktabs}
\usepackage{multirow}
\usepackage{setspace}
\usepackage{subcaption}

\newcommand{\md}{\mathrm{d}}
\newcommand{\mN}{\mathcal{N}}
\newcommand{\mA}{\mathcal{A}}
\newcommand{\mM}{\mathcal{M}}
\newcommand{\mH}{\mathcal{H}}
\newcommand{\mU}{\mathcal{U}}

\DeclareMathOperator{\mE}{\mathbb{E}}

\DeclareMathOperator{\Tr}{Tr}
\DeclareMathOperator*{\argmin}{argmin}
\newcommand{\trans}{\mathsf{T}}

\usepackage[dvipsnames,svgnames,x11names]{xcolor}
\makeatletter
\def\ps@pprintTitle{%
 \let\@oddhead\@empty
 \let\@evenhead\@empty
 \def\@oddfoot{\centerline{\thepage}}%
 \let\@evenfoot\@oddfoot}
\makeatother

\begin{document}

\begin{frontmatter}



\title{State-observation augmented diffusion model for nonlinear assimilation with unknown dynamics}


\author[pku]{Zhuoyuan Li} %% Author name
\ead{zy.li@alumni.pku.edu.cn}
\author[bicmr,cmlr]{Bin Dong\corref{cor1}} 
\ead{dongbin@math.pku.edu.cn}
\author[whu,pku]{Pingwen Zhang\corref{cor1}}
\ead{pzhang@pku.edu.cn}

\cortext[cor1]{Co-corresponding authors.}

\affiliation[pku]{
organization={School of Mathematical Sciences, Peking University},
city={Beijing},
postcode={100871},
state={Beijing},
country={China}
}
\affiliation[bicmr]{
organization={Beijing International Center for Mathematical Research and the New Cornerstone Science Laboratory, Peking University},
city={Beijing},
postcode={100871},
state={Beijing},
country={China}
}
\affiliation[cmlr]{
organization={Center for Machine Learning Research, Peking University},
city={Beijing},
postcode={100871},
state={Beijing},
country={China}
}
\affiliation[whu]{
organization={School of Mathematics and Statistics, Wuhan University},
city={Wuhan},
postcode={430072},
state={Hubei},
country={China}
}

% \affiliation{organization={},%Department and Organization

%% Abstract
\begin{abstract}
    Data assimilation has become a key technique for combining physical models with observational data to estimate state variables. However, classical assimilation algorithms often struggle with the high nonlinearity present in both physical and observational models. To address this challenge, a novel generative model, termed the State-Observation Augmented Diffusion (SOAD) model is proposed for data-driven assimilation.
    The marginal posterior associated with SOAD has been derived and then proved to match the true posterior distribution under mild assumptions, suggesting its theoretical advantages over previous score-based approaches. Experimental results also indicate that SOAD may offer improved performance compared to existing data-driven methods.
\end{abstract}

% \includegraphics[width=\textwidth]{SOADarchitecture.pdf}

% \begin{highlights}
% \item State-observation augmented dynamics help relax linear assumptions
% % \item Simplifies assimilation with an equivalent linear structure, relaxing assumptions
% \item Real posterior can be achieved under mild theoretical assumptions
% \item Forward-diffusion corrector improves reverse-time stability
% \item Experiments confirm effectiveness with varied observational operators
% \end{highlights}

%% Keywords
% \begin{keyword}
% %% keywords here, in the form: keyword \sep keyword
% data assimilation \sep score-based generative model \sep Bayesian inference
% %% PACS codes here, in the form: \PACS code \sep code
% % TODO
% %% MSC codes here, in the form: \MSC code \sep code
% %% or \MSC[2008] code \sep code (2000 is the default)

% \end{keyword}

\end{frontmatter}



% \linenumbers
\section{Introduction}
The advent of modern observational facilities and devices has led to an exponential increase in the volume of data gathered from the physical world. Consequently, data assimilation (DA), as an uncertainty quantification technique fusing observations with physical models, has attracted significant attention. Over the past few decades, the DA community has developed a series of methods to tackle the assimilation problem under various assumptions and constraints. From early approaches such as the nudging method and optimal interpolation to modern techniques including the Kalman filter (KF) \cite{kalman1960} and its variants \cite{jazwinski2007stochastic,LETKF}, variational methods, and their hybrid versions \cite{En4DVar,4DEnVar}, researchers have demonstrated both theoretical and practical effectiveness. Nonetheless, the high dimensionality and intrinsic nonlinearity of physical and observational models continue to pose substantial challenges.

In this work, we follow the four-dimensional variational data assimilation (4D-Var) formulation, which aims to estimate the posterior distribution of state variables over a fixed temporal window given a sequence of observations.
Mathematically, consider a dynamical system described by
\begin{equation}\label{eq:discrete-dynamics}
    \begin{cases}
        \bm x_k=\mM_k(\bm x_{k-1},\bm\eta_k),\quad\bm\eta_k\sim\mathscr{D}_k^X, \\
        \bm y_k=\mH_k(\bm x_k)+{\bm\epsilon_k},\quad{\bm\epsilon_k}\sim\mathscr{D}_k^Y,
    \end{cases}
\end{equation}
where $\bm x_k$ and $\bm y_k$ denote the state and observation variables at time step $k$, respectively.
The observation noise ${\bm\epsilon_k}\sim\mathscr D_k^Y$ captures uncertainties in the observational data, while the term $\bm\eta_k\sim\mathscr D_k^X$ accounts for possible model errors in the physical model $\mM_k$.
The goal is to estimate the posterior distribution $p(\bm x_{S:T}\mid\bm y_{S:T})$ for the state variables over the interval from $S$ to $T$, i.e., we use the abbreviation $\bm x_{S:T}$ for $\{\bm x_k\}_{k=S}^T$.
By Bayes' rule, the posterior of the states given the observations can be expressed as
\begin{equation}\label{eq:posterior-bayes}
    p(\bm x_{S:T}\mid\bm y_{S:T}) =\frac{p(\bm y_{S:T}\mid\bm x_{S:T})p(\bm x_{S:T})}{p(\bm y_{S:T})} \propto p(\bm x_{S:T})p(\bm y_{S:T}\mid\bm x_{S:T}).
\end{equation}
Classical variational methods \cite{Asch2016,Evensen2022} further expand \cref{eq:posterior-bayes} under independence assumptions for the observational and model errors, yielding
\begin{equation}
    p(\bm x_{S:T}\mid\bm y_{S:T}) \propto p(\bm x_S)\prod_{k=S}^{T-1}p(\bm x_{k+1}\mid\bm x_k)\prod_{k=S}^Tp(\bm y_k\mid\bm x_k)
\end{equation}
A common approach then involves \textit{maximum a posteriori} (MAP) estimation under Gaussian assumptions and solves
\begin{equation}
\resizebox{.93\hsize}{!}{$
    \bm x_{S:T}^*=\argmin\limits_{\bm x_{S:T}}\left\{\frac12\left\|\bm x_S-\bm x_S^\mathrm{b}\right\|_{\bm B}^2+\frac12\sum\limits_{k=S}^{T-1}\left\|\bm x_{k+1}-\mM_k^X(\bm x_k)\right\|_{\bm Q_k}^2+\frac12\sum\limits_{k=S}^T\left\|\bm y_k-\mH_k(\bm x_k)\right\|_{\bm R_k}^2\right\}
    $}
\end{equation}
by the variational principle, where
\begin{equation}
    \bm x_S\sim\mathcal N(\bm x_S^\mathrm{b},\bm B),\quad\bm\eta_k\sim\mathcal N(\bm0,\bm Q_k),\quad{\bm\epsilon_k}\sim\mathcal N(\bm0,\bm R_k),
\end{equation}
and the model error $\bm\eta_k$ is assumed additive. Here, we use the notation $\|\bm a\|_{\bm A}^2:=\bm a^\trans\bm A^{-1}\bm a$ for simplicity.

Despite the success of classical variational data assimilation methods, they still face several fundamental limitations, primarily in their treatment of uncertainty \cite{Bannister2017,Evensen2022}. The background-error covariance is typically prescribed and assumed to follow a static Gaussian distribution, which does not capture the evolving, nonlinear, and flow-dependent nature of real atmospheric uncertainties. This mismatch can lead to suboptimal weighting of observations and errors in state estimation.
Additionally, the iterative minimization process requires repeated evaluations of both the forward physical model and its adjoint, making large-scale applications computationally expensive \cite{rabier2003variational}.
Moreover, deriving and maintaining tangent linear models and adjoint models adds further complexity, restricting the flexibility and applicability of variational methods.
These challenges highlight the need for alternative approaches that can better represent uncertainty and efficiently handle nonlinear, high-dimensional systems.

As noted by \cite{Schultz2021DLvsNWP,Geer2021}, the rapid influx of observational data, such as satellite observations and in-situ scientific measurements, has encouraged researchers to explore data-driven approaches. Machine learning, and in particular deep learning (DL), has shown its remarkable capability in learning nonlinear correlations and extracting high-dimensional features from training data. Inspired by the success of DL in fields such as computer vision (CV) \cite{DL4CV-review,kirillov2023segment} and natural language processing (NLP) \cite{devlin2018bert,touvron2023llama,zhao2023LLMsurvey}, scientists are increasingly applying DL techniques within the earth and atmospheric sciences community. Recent success in training large foundation models for weather forecasting \cite{chen2023foundation,FourCastNet,Pangu,GraphCast,FuXi}, have demonstrated the potential of data-driven models in handling complex global atmospheric physics.
These advancements indicate a shift towards data-driven methodologies, which may enhance our understanding and predictive capabilities in atmospheric sciences.

However, unlike forecasting or nowcasting tasks that can be directly handled by common autoregressive models such as RNNs \cite{LSTM,cho2014propertiesGRU,shi2015convolutional} and Transformers \cite{vaswani2017attention,liu2021swin}, the assimilation problem is inherently more complex. The core of assimilation involves merging observations into the physical prior, which necessitates a more sophisticated model to handle the intricate interactions between physical models and observations. Furthermore, the assimilation problem is essentially a Bayesian inference task, which requires the estimation of the posterior distribution of the state variables rather than just a deterministic prediction for future states, and due to the high dimensionality and nonlinearity of the physical and observational models, the posterior distribution is often intractable.

Motivated by the success of DL, we seek to reframe the assimilation problem through the lens of deep generative models (DGMs) \cite{Ruthotto2021DGMintro,Sam2022DGMreview}, which have demonstrated strong capabilities in capturing complex data distributions.
The connection between assimilation and DGMs is intuitive: assimilation aims to estimate the posterior distribution of state variables by integrating prior knowledge from physical models with observational data, while DGMs also provide data distributions, but by training with large datasets.
In assimilation, physical models contribute to the prior, which is calibrated by the likelihood derived from observational operators and data.
Once the posterior is obtained, key statistical quantities such as the mean field and uncertainty estimates can be inferred.
DGMs, on the other hand, rely on large datasets sampled from the target distribution, assuming that all hidden patterns can be extracted without explicit physical knowledge. By incorporating learned priors from DGMs with suitable likelihood modeling, we introduce an alternative approach to assimilation, particularly beneficial when the underlying physics are not fully known.

The rest of the paper is organized as follows.
First, we review the related works on data assimilation and deep generative models in \cref{sec:related-works}. Next, in \cref{sec:methodology}, we present the mathematical formulation of the assimilation task we focus on. Some basic concepts of conditional score-based generative models are introduced, followed by a detailed deduction of our SOAD model and its theoretical analysis. Finally, experiments on a two-layer quasi-geostrophic model will be exhibited in \cref{sec:experiments} to demonstrate the effectiveness of our SOAD model, and \cref{sec:conclusion} concludes the paper with a summary and future work.

% section 2
\section{Related works and contributions}\label{sec:related-works}
\subsection{Data assimilation frameworks}
Data assimilation has been studied for decades. 
{In general, data assimilation can be categorized into \textit{filtering} and \textit{smoothing}. Filtering aims to estimate the current state given observations up to the current time, while smoothing involves estimating states using both past and future observations. The choice between the two types depends on not only the specific application and objective of the assimilation task but also the availability of data and computational resources.}

{Ensemble Kalman filter (EnKF) and its variants are representative classical methods for the filtering problem, which adopt Gaussian assumptions and use an ensemble of samples to represent the posterior's mean and covariance to seek a balance between accurate probabilistic modeling and computational efficiency. In contrast, the 3D-Var method solves for the MAP estimate for the posterior, providing a ``best-fit'' solution.
}
Particle filters, specially designed for {the} nonlinear problem, approximate the posterior by a weighted sum of Dirac deltas, allowing non-Gaussian distributions. 
Despite recent progress in the particle flow filter (PFF) \cite{Daum2013PFF,Daum2018PFF} and its demonstrated potential in high-dimensional settings \cite{Hu2021PFF-high-dimensional}, {the} need for a large number of particles remains a computational bottleneck. 

{To address the smoothing problem, one of the most intuitive approaches is the Kalman smoother (KS), which includes an additional backward pass to refine estimates using future information compared to the original KF. As ensemble-based extensions of KS, Ensemble Kalman Smoother (EnKS) \mbox{\cite{Evensen2000EnKS}} and iterative EnKS (IEnKS) \mbox{\cite{Bocquet2014IEnKS}} utilize an ensemble forward propagation to obtain better convergence for nonlinear problems.
4D-Var formulates a variational problem over a time window to retrieve the most likely trajectory, making it inherently a smoothing approach as well.}

All of these classical methods rely on physical knowledge and can be prohibitively expensive in high-dimensional problems that require repeated ensemble updates or large-scale optimizations. Despite these challenges, they remain indispensable for reliably integrating observations with complex dynamical models in many scientific and engineering applications.
Nowadays, deep learning (DL)-enhanced data assimilation has gained increasing attention \cite{Cheng2023MLDAUQ}.
A pioneering example is the framework of \cite{Brajard2021}, which replaces physical models with neural networks, demonstrating the feasibility of data-driven assimilation.
Leveraging the flexibility and fast inference of neural networks, some studies have proposed using them as fast emulators \cite{Chattopadhyay2023H-EnKF} or correctors \cite{Hammoud2024RL-DA} for physical models in hybrid assimilation frameworks built on classical algorithms.
Another research direction focuses on addressing the high-dimensionality challenge through latent assimilation (LA) \cite{amendola2020data}, which seeks a lower-dimensional representation of the state variables to be assimilated using either linear Reduced-Order Models (ROMs) \cite{Pawar2022NIROM-DA,Cheng2023GLA} or neural networks \cite{Peyron2021LAwithAE,LatentspaceDA-RNN}.  Although methods like LAINR \cite{LAINR2024} extend the flexibility of LA, they still rely on well-established classical assimilation algorithms as backbones to operate in the latent space.

To better capture non-Gaussian posteriors, recent work has explored the Score-based Filter (SF) \cite{Bao2024SF} and its variants, such as the Ensemble Score-based Filter (EnSF) \cite{Bao2024EnSF} and Latent-EnSF \cite{Si2024Latent-EnSF}, which rely on diffusion probabilistic models (DPMs) for posterior estimation. These methods have shown promise for nonlinear assimilation tasks but typically assume availability of a physical model.
More recently, Score-based Data Assimilation (SDA) \cite{Rozet2023sda,Rozet2023sda-2lqg} has been introduced to integrate physical models into the learning of background priors within a variational assimilation framework, producing promising results for subsampled observations even without any explicit knowledge of physical models. However, as demonstrated in our experiments, SDA still struggles with nonlinear observations.

{In the modern numerical weather prediction pipeline, increasing attention has been given to smoothing-based methods, particularly 4D-Var and its variants \mbox{\cite{Clayton2013,Bonavita2016,Lorenc2015,Zhang2012E4DVar}}, due to their ability to incorporate time-evolving observations and improve the dynamical consistency of the atmospheric state \mbox{\cite{Andersson1994,Thepaut1996}}. These advances highlight a growing need for learning-based smoothing algorithms that can effectively assimilate observations over a time window and handle the non-Gaussianity arising from nonlinear dynamics and observation operators. Motivated by this, our proposed method adopts a smoothing perspective and seeks to estimate latent trajectories conditioned on both past and future observations, with a focus on addressing challenges related to nonlinearity and posterior complexity.}
\Cref{tab:comparison-DA-framework} summarizes key features of some typical data assimilation approaches. Bolded entries highlight capabilities that surpass those of other methods in the same category.

\begin{table}
    \centering
    \resizebox{\columnwidth}{!}{%
    \begin{tabular}{l|c|c|c}
        \toprule
        Method                                                                       & physical model                                    & posterior modeling                    & comput. cost                        \\
        \midrule
        KF\cite{kalman1960}, EnKF\cite{jazwinski2007stochastic}, LETKF\cite{LETKF}            & required                                          & mean \& covariance                    & high                                      \\
        PF, PFF\cite{Daum2013PFF,Daum2018PFF,Hu2021PFF-high-dimensional}                      & required                                          & weighted Dirac sum                    & very high                                 \\
                        SF\cite{Bao2024SF}                                                                    & required                                          & \textbf{generative model} & high                                      \\
                EnSF\cite{Bao2024EnSF}, Latent-EnSF\cite{Si2024Latent-EnSF}                           & required                                          & \textbf{generative model} & \textbf{low}                  \\
        \midrule
        % \multirow{2}{*}{LA\cite{Peyron2021LAwithAE,Pawar2022NIROM-DA,Cheng2023GLA,LAINR2024}} & \multirow{2}{*}{\textbf{data-driven}} & same as                               & \multirow{2}{*}{\textbf{low}} \\
                                                                                              % &                                                   & the DA backbone                       &                                           \\
        % \midrule
        % \midrule
        % SSLS\cite{Ding2024SSLS}                                               & required                         & \underline{\textbf{MCMC}}       & high                     \\
                        4D-Var                                                                             & required                                          & MAP                                   & high                                      \\
                KS, EnKS\cite{Evensen2000EnKS}, IEnKS\cite{Bocquet2014IEnKS}            & required                                          & mean \& covariance                    & high                                      \\
        SDA\cite{Rozet2023sda,Rozet2023sda-2lqg}, SOAD (ours)                                 & \textbf{data-driven}                  & \textbf{generative model} & \textbf{low}                  \\
        \bottomrule
    \end{tabular}%
    }
    \caption{{Comparison of some typical filtering and smoothing approaches.}}
    \label{tab:comparison-DA-framework}
\end{table}

\subsection{Deep generative models}
Deep generative models (DGMs) have been consistently studied in the field of deep learning.
As one of the earliest DGMs, variational AutoEncoders (VAEs) \cite{kingma2013VAE,doersch2016vae-tutorial} learn the target data distribution by maximizing the evidence lower bound (ELBO).
Generative Adversarial Networks (GANs) \cite{Goodfellow2020GAN,Creswell2018GAN-overview} are another kind of DGMs, which learn the distribution through a zero-sum game between generators and discriminators.
Alternatively, Normalizing Flows (NFs) \cite{dinh2014nice,rezende2015variational,kobyzev2020normalizing} are employed not only for generating samples but also for explicitly evaluating probability densities by ensuring each layer to be invertible at the expense of flexibility of network architectures.

{
Inspired by \mbox{\cite{sonderby2016ladder}}, the most recent diffusion probabilistic models (DPMs) \mbox{\cite{luo2022understanding,yang2023diffusion,yang2024survey}} have achieved impressive performances across numerous data generation and restoration tasks, particularly in computer vision and natural language processing domains.
Specifically, DPMs have shown state-of-the-art performances in image generation and restoration tasks such as denoising, inpainting, and super-resolution \mbox{\cite{Ho2020DDPM,song2020DDIM,rombach2022latent-diffusion}}, as well as in text-to-image and video generation \mbox{\cite{singer2022make-a-video}}. These tasks often involve linear measurement operators (e.g., masking, blurring, or downsampling), and thus the associated inverse problems are typically linear, enabling efficient deployment of diffusion models through techniques like iterative reconstruction guided by the learned score function or posterior sampling \cite{daras2024survey}.
}

{
More recently, diffusion models have been adapted to tackle complex inverse problems in scientific computing. Unlike many standard image restoration scenarios, scientific computing tasks such as phase retrieval \mbox{\cite{Chung2023DPS,Li2024DiffFPR}}, seismic inversion \mbox{\cite{Wang2024seismic,baldassari2023seismic,wang2023prior}}, and topology optimization \mbox{\cite{bastek2023inverse,maze2023diffusion,giannone2023aligning}} generally involve nonlinear, high-dimensional, and often ill-posed inverse problems. In these contexts, DPMs are leveraged to model the prior distribution of states efficiently and sample from complex posterior distributions conditioned on nonlinear observations.
}

\subsection{Main contributions}
In this study, we propose a novel data-driven deep learning (DL) approach specifically designed for posterior estimation to handle the assimilation problem.
To address the challenges of nonlinearity, we introduce our State-Observation Augmented Diffusion (SOAD) model, which relaxes the linear constraints typically imposed on physical and observational models in most existing assimilation frameworks. Similar to previous DL-based assimilation methods, our SOAD model is fully data-driven and does not require any prior knowledge of the physical models. 
{Importantly, while our network training still relies on observational data that implicitly reflects the observation process, our model does not require an explicit physical formulation of the observational operator. That is, as long as paired state-observation samples are available, SOAD can be trained and subsequently perform assimilation with observations from multiple heterogeneous sources without requiring any knowledge of how these observations are physically related to the state variables.}
The flexibility and adaptability make our approach suitable for the real-world applications. Our main contributions are as follows:
\begin{itemize}
    \item We propose a novel state-observation augmented diffusion model specially designed for nonlinear assimilation tasks. By introducing the state-observation augmented structure, we establish an equivalent linear form of the original assimilation problem, relaxing the linearity assumptions for both the physical and observational models.
    \item We derive the marginal posterior distribution of the state variables associated with our SOAD model and show that it matches the real posterior under mild assumptions, indicating the theoretical superiority of our SOAD model over previous score-based assimilation approaches.
    \item A procedure named forward-diffusion corrector is introduced to stabilize the reverse-time generation process of the diffusion model, providing corrections matching the real distribution especially under Gaussian noise.
    \item Experiments with a two-layer quasi-geostrophic model involving various observational operators are conducted to demonstrate the effectiveness of our SOAD model compared to previous works.
\end{itemize}

% section 3
\section{Methodology}\label{sec:methodology}
\subsection{Problem settings and assumptions}\label{sec:problem-settings-assumptions}
Recall that we consider a discrete dynamical system with an observational model formulated as
\begin{equation}\label{eq:discrete_dynamical_system}
    \begin{cases}
        \bm x_k=\mM_k^X(\bm x_{k-1},\bm\eta_k),\quad\bm\eta_k\sim\mathscr{D}_k^X, \\
        \bm y_k=\mH_k(\bm x_k)+{\bm\epsilon_k},\quad{\bm\epsilon_k}\sim\mathscr{D}_k^Y,
    \end{cases}
\end{equation}
where the additional superscript ``$X$'' indicates that the model $\mM_k^X$ acts in the physical space for $\bm x_k$.
We follow the setup described in \cite{Rozet2023sda,Song2021scorebased}, assuming access to a sufficiently large dataset of physical states corresponding to an unknown physical model $\mM_k^X$. {The noise distributions $\mathscr{D}_k^Y$ associated with all observational operators are assumed to be known as well.}
Given a sequence of observations $\bm y_{S:T}$, our objective is to effectively estimate the posterior distribution of the state variables $\bm x_{S:T}$, denoted by $p\left(\bm x_{S:T}\mid\bm y_{S:T}\right)$.

To keep our approach general, we impose only the following assumptions:
\begin{itemize}
    \item \textbf{Time-independence or periodicity:} The physical model $\mM_k^X$ is either constant over time or exhibits periodic behavior.
    \item \textbf{Decomposable observational operators:} Although the observational operator $\mH_k$ may vary with time, there exists a time-independent operator $\mH$ such that $\mH_k=\bm S_k\circ\mH$ for some time-varying linear operator $\bm S_k${ with orthogonal rows}.
\end{itemize}
{
These assumptions are pragmatic in many practical contexts. For instance, numerical weather prediction frequently exhibits daily or seasonal periodicity in the underlying physical model $\mM_k^X$. The first assumption ensures the continued applicability and effectiveness of data-driven methods trained on historical data, a standard practice in deep learning-based assimilation methodologies \mbox{\cite{Rozet2023sda,Chattopadhyay2023H-EnKF,Peyron2021LAwithAE,Pawar2022NIROM-DA,LAINR2024}}. The observational operator $\mathcal{H}_k$ typically varies over time due to observational constraints or infrastructure limitations \cite{lahoz2010data}, but the range of possible observational configurations usually remains fixed. Therefore, we may set
}
% {
% We make the first assumption to ensure data-driven methods remain valid after training on sufficient historical data. The second assumption is crucial for our theoretical deduction in later sections.
% These assumptions are pragmatic in many practical contexts. For instance, numerical weather prediction often assumes daily or seasonal periodicity \cite{satoh2013atmospheric} for the physical model $\mM_k^X$. The observational operator $\mH_k$ may vary at different time steps due to limitations of observational facilities, yet the set of all possible observational operators typically does not change. 
% Hence, we may set
% }
\begin{equation}
    \mH(\bm x_k)=\begin{pmatrix}
        \mH^{(1)}(\bm x_k) \\
        \mH^{(2)}(\bm x_k) \\
        \vdots             \\
        \mH^{(K)}(\bm x_k)
    \end{pmatrix}
\end{equation}
as a concatenation of all possible observational outputs, where each $\mH^{(i)}$ corresponds to one type of observation measured on full domain for $i=1,2,\cdots, K$. This suggests that the variation in observational operators for different time steps can be adequately captured by a {``subsampling matrix''} $\bm S_k$ acting on $\mH(\bm x_k)${. Here, a subsampling matrix is a binary matrix that selects a subset of elements (or locations) from the full state vector.}
% {, which means $\bm S_k$ is a row-orthogonal matrix with all the entries being either $0$ or $1$}. 
While traditional assimilation often presume Gaussian noise distributions $\mathscr{D}^X$ and $\mathscr{D}^Y$, we will demonstrate later that these assumptions can be relaxed as well.

\subsection{Score-based data assimilation}
Score-based data assimilation is inspired by score-based generative models \cite{Song2021scorebased}, a widely studied class of diffusion probabilistic models (DPMs) in machine learning. These models offer an alternative perspective by leveraging stochastic differential equations for data generation. In this section, we briefly review the fundamental concepts underlying score-based generative models and how they can be applied to the data assimilation problem.

\subsubsection{Score-based generative models}\label{sec:score-based-generative-models}
Let $\bm x\sim p_\mathrm{data}(\bm x)$ denote the data distribution we aim to learn. By progressively adding Gaussian noise, the distribution undergoes an approximate transformation into a standard normal distribution, and the whole process can be modeled as a forward-time diffusion process. Mathematically, following the notation in \cite{Song2021scorebased}, we describe the diffusion process as a linear stochastic differential equation (SDE) formulated as
\begin{equation}\label{eq:forward_sde}
    \md\bm x_t=f(t)\bm x_t\md t+g(t)\md\bm w_t,\quad t\in[0,1],\quad \bm x_0\sim p_\mathrm{data}(\bm x_0),
\end{equation}
where $f(t)$ and $g(t)$ represent the drift term and diffusion coefficient, respectively, and $\bm w_t$ denotes the standard Wiener process. The transition kernel from {$\bm x_0$} to $\bm x_t$ is then given by
\begin{equation}\label{eq:transition-kernel}
    p_{t\mid0}(\bm x_t\mid \bm x_0)=\mN\left(\bm x_t;\mu_t\bm x_0,\sigma_t^2\bm I\right)
\end{equation}
with the mean $\mu_t$ and the variance $\sigma_t^2$ explicitly \cite{Sarkka2019} given by
\begin{equation}
    \mu_t=\exp\left(\int_0^t f(s)\md s\right),\quad\sigma_t^2=\int_0^t\exp\left(2\int_s^t f(u)\md u\right)g(s)^2\md s.
\end{equation}
Define the marginal distribution for time $t$ as $p_t(\bm x)$.
With appropriate choices of $f(t)$ and $g(t)$ so that $\mu_1=\sigma_0=0$ and $\mu_0=\sigma_1=1$, the forward diffusion process \cref{eq:forward_sde} transforms the data distribution into a standard Gaussian marginal $p_1(\bm x)=\mN(0,\bm I)$. Under such circumstances, the reverse-time evolution, characterized by
\begin{equation}
    \md\bm x_t=\left[f(t)-g(t)^2\nabla_{\bm x_t}\log p_t(\bm x_t)\right]\md t+g(t)\md{\bm w}_t, %\cev
\end{equation}
is also a diffusion process \cite{Anderson1982}, where the gradient term $\nabla_{\bm x_t}\log p_t(\bm x_t)$, known as the score function, plays a crucial role in guiding the reverse diffusion. {With a slight abuse of notation}, we still let $\bm w_t$ denote the reverse-time Wiener process.
Consequently, as long as a good estimation of the score function is obtained for all $t$, we may first sample an initial state $\bm x_1\sim\mN(0,\bm I)$ and then simulate the reverse-time process starting from $\bm x_1$ to generate a sample $\bm x_0$ that approximately follow the distribution $p_\mathrm{data}(\bm x)$.
% \deleted[id=rev2]{
% Classical numerical solvers for general SDEs, such as the Euler-Maruyama and stochastic Runge-Kutta methods, as well as ancestral sampling \mbox{\cite{Ho2020DDPM,Song2021scorebased}}, are applicable. Recently, to improve sampling efficiency, the exponential integrator (EI) discretization scheme \mbox{\cite{Zhang2022FastSampling}}
% }
% \begin{equation}\label{eq:EI-predictor}
% \deleted[id=rev2]{
%     \bm x_{t_-}\leftarrow\frac{\mu_{t_-}}{\mu_t}\bm x_t+\left(\frac{\sigma_{t_-}}{\sigma_t}-\frac{\mu_{t_-}}{\mu_t}\right)\sigma_t\bm\varepsilon_\theta(\bm x_t,t)
%     }
% \end{equation}
% \deleted[id=rev2]{
% has also been proposed, where the subscript ``$t_-$'' denotes the next time step. These schemes often work together with the predictor-corrector sampling strategy by performing a few steps of Langevin Monte Carlo (LMC) sampling \mbox{\cite{Song2021scorebased}}
% }
% \begin{equation}\label{eq:LMC-corrector}
% \deleted[id=rev2]{
%     \bm x_t\leftarrow\bm x_t+\frac{\delta}2\bm s_\theta(\bm x_t,t)+\sqrt\delta\bm z,\quad\bm z\sim\mN(0,\bm I)
%     }
% \end{equation}
% \deleted[id=rev2]{
% for better performances.
% }

To develop an effective score-based generative model, intuitively one may minimize the Fisher information distance (\cite{Dasgupta2008asymptotic}, Definition 2.5)
\begin{equation}
    d(\bm\theta,t)=\mE_{\bm x_t\sim p_t}\left\|\bm s_{\bm\theta}(\bm x_t,t)-\nabla_{\bm x_t}\log p_t(\bm x_t)\right\|^2,
\end{equation}
where
\begin{equation}
    p_t(\bm x)=\int p_{t\mid0}(\bm x\mid\bm x_0)p_\mathrm{data}(\bm x_0)\md\bm x_0
\end{equation}
is the marginal density and $\bm s_{\bm\theta}(\bm x_t,t)$ is a surrogate model for the score function implemented by a neural network with parameters $\bm\theta$. By assuming some weak regularity conditions \cite{Hyvarinen2005}, we have
\begin{equation}
    \begin{aligned}
        d(\bm\theta,t) & =\mE_{\bm x_t\sim p_t}\left\|\bm s_{\bm\theta}(\bm x_t,t)-\nabla_{\bm x_t}\log p_t(\bm x_t)\right\|^2                                                                        \\
                       & =\frac12\mE_{\bm x_t\sim p_t}{\left[\left\|\bm s_{\bm\theta}(\bm x_t,t)\right\|^2+\Tr\nabla_{\bm x_t}\bm s_{\bm\theta}(\bm x_t,t)\right]              }           \\
                       & =\frac12\mE_{\bm x_0\sim p_0}\mE_{\bm x_t\sim p_{t\mid0}(\cdot\mid\bm x_0)}{\left[\left\|\bm s_{\bm\theta}(\bm x_t,t)\right\|^2+\Tr\nabla_{\bm x_t}\bm s_{\bm\theta}(\bm x_t,t)\right]}     \\
                       & =\frac12\mE_{\bm x_0\sim p_0}\mE_{\bm x_t\sim p_{t\mid0}(\cdot\mid\bm x_0)}\left\|\bm s_{\bm\theta}(\bm x_t,t)-\nabla_{\bm x_t}\log p_{t\mid0}(\bm x_t\mid\bm x_0)\right\|^2 \\
                       & =\frac12\mE_{\bm x_0\sim p_0}\mE_{\bm\varepsilon\sim\mN(\bm0,\bm I)}\left\|\bm s_{\bm\theta}(\mu_t\bm x_0+\sigma_t\bm\varepsilon,t)+\sigma_t^{-1}\bm\varepsilon\right\|^2
    \end{aligned}
\end{equation}
as discussed in \cite{Vincent2011connection}.
In practice, $\bm s_\theta(\bm x_t,t)$ is usually reparameterized as
\begin{equation}\label{eq:epsilon-and-score}
    \bm\varepsilon_\theta(\bm x_t,t)=-\sigma_t\bm s_\theta(\bm x_t,t),
\end{equation}
and we consider minimizing
\begin{equation}\label{eq:denoising-loss}
    L(\bm\theta)=\mE_{t\sim\mU_{[0,1]}}\sigma_t^2d(\bm\theta,t)=\frac12\mE_{\substack{t\sim\mU_{[0,1]}\\\bm\varepsilon\sim\mN(0,\bm I)\\\bm x_0\sim p_0(\bm x)}}\left\|\bm\varepsilon_\theta(\mu_t\bm x_0+\sigma_t\bm\varepsilon,t)-\bm\varepsilon\right\|^2.
\end{equation}

{
Once a good estimator for the score funtion is obtained, it suffices to evolve the backward diffusion process by discretizing the time steps.
}
{
Classical numerical solvers for general SDEs, such as the Euler-Maruyama and stochastic Runge-Kutta methods, as well as ancestral sampling \mbox{\cite{Ho2020DDPM,Song2021scorebased}}, are applicable. Recently, to improve sampling efficiency, the exponential integrator (EI) discretization scheme \mbox{\cite{Zhang2022FastSampling}}
}
\begin{equation}\label{eq:EI-predictor}
    {
    \bm x_{t_-}\leftarrow\frac{\mu_{t_-}}{\mu_t}\bm x_t+\left(\frac{\sigma_{t_-}}{\sigma_t}-\frac{\mu_{t_-}}{\mu_t}\right)\sigma_t\bm\varepsilon_\theta(\bm x_t,t)
    }
\end{equation}
{
has also been proposed, where the subscript ``$t_-$'' denotes the next time step. These schemes often work together with the predictor-corrector sampling strategy by performing a few steps of Langevin Monte Carlo (LMC) sampling \mbox{\cite{Song2021scorebased}}
}
\begin{equation}\label{eq:LMC-corrector}
{
    \bm x_t\leftarrow\bm x_t+\frac{\delta}2\bm s_\theta(\bm x_t,t)+\sqrt\delta\bm z,\quad\bm z\sim\mN(0,\bm I)
    }
\end{equation}
{
for better performances, where we let $\delta$ denote the time step.
}

\subsubsection{Conditional score models}
While score-based generative models we have introduced above can be directly used for unconditional sampling with well-trained denoising models, assimilation tasks require estimating the posterior distribution $p(\bm x\mid\bm y)$, where we omit the subscript ``$S:T$'' for simplicity. Analogous to the unconditional case, the posterior distribution can be approximated with the reverse-time diffusion process defined as
\begin{equation}
    \md\bm x_t=\left[f(t)-g(t)^2\nabla_{\bm x_t}\log p_t(\bm x_t\mid\bm y)\right]\md t+g(t)\md{\bm w}_t,\quad\bm x_1\sim\mN(\bm0,\bm I),
\end{equation}
where the conditional score function $\nabla_{\bm x_t}\log p_t(\bm x_t\mid\bm y)$ corresponding to the marginal posterior
\begin{equation}\label{eq:marginal-posterior}
    p_t(\bm x_t\mid\bm y)=\int p_{t\mid0}(\bm x_t\mid\bm x)p(\bm x\mid\bm y)\md\bm x
\end{equation}
needs to be learned.
Note that the conditional score function can be decomposed as
\begin{equation}\label{eq:conditional-score-bayes}
    \begin{aligned}
        \nabla_{\bm x_t}\log p_t(\bm x_t\mid\bm y) & =\nabla_{\bm x_t}\log\int p_{t\mid0}(\bm x_t\mid\bm x)p(\bm x\mid\bm y)\md\bm x                                    \\
                                                   & =\nabla_{\bm x_t}\log\int \frac{p_{t\mid0}(\bm x_t\mid\bm x)p(\bm y\mid\bm x)p(\bm x)}{p(\bm y)}\md\bm x           \\
                                                   & =\nabla_{\bm x_t}\log\int \frac{p_{0\mid t}(\bm x\mid\bm x_t)p(\bm y\mid\bm x)p_t(\bm x_t)}{p(\bm y)}\md\bm x      \\
                                                   & =\nabla_{\bm x_t}\log p_t(\bm x_t)+\nabla_{\bm x_t}\log\int p(\bm y\mid\bm x)p_{0\mid t}(\bm x\mid\bm x_t)\md\bm x
    \end{aligned}
\end{equation}
by Bayes' formula, where the additional term
\begin{equation}\label{eq:conditional-likelihood}
    \nabla_{\bm x_t}\log p_t(\bm y\mid\bm x_t)=\nabla_{\bm x_t}\log\int p(\bm y\mid\bm x)p_{0\mid t}(\bm x\mid\bm x_t)\md\bm x
\end{equation}
is usually called the adversarial gradient.

To incorporate conditional information, one may introduce a separate network for the adversarial gradient alongside the original denoising network as suggested in \cite{dhariwal2021diffusion,ho2022classifier-free}. Alternatively, a conditional denoising network $\bm\varepsilon_\theta(\bm x_t,t;\bm y)$ can be trained directly to approximate $-\sigma_t\nabla_{\bm x_t}\log p_t(\bm x_t\mid\bm y)$, and such an idea has been widely used in many class-conditional generation and text-to-image/video generation tasks \cite{rombach2022latent-diffusion,singer2022make-a-video}.

Instead of training a different denoising network, we focus on estimating the reverse-time transition kernel $p_{0\mid t}$ in \cref{eq:conditional-likelihood} to compute the term $p_t(\bm y\mid\bm x_t)$ explicitly. By Tweedie's formula, DPS \cite{Chung2023DPS} proposes a delta-function approximation for the kernel $p_{0\mid t}$ centered at
\begin{equation}
    \hat{\bm x}_0(\bm x_t,t)=\mE[\bm x_0\mid\bm x_t]=\mu_t^{-1}[\bm x_t+\sigma_t^2\nabla_{\bm x_t}\log p_t(\bm x_t)].
\end{equation}
Based on the uninformative-prior assumption, DMPS \cite{Meng2022DMPS} replaces the kernel $p_{0\mid t}$ with a Gaussian distribution, but it is only suitable for linear observational models. SDA \cite{Rozet2023sda,Rozet2023sda-2lqg} suggests a balance between flexibility and accuracy by introducing an auxiliary hyperparameter $\gamma$ in the modeling of the kernel $p_{0\mid t}$. Nonetheless, such an approximation is still based on the linearity assumption for the observational operator, which has shown limitations in our experiments. Take a nonlinear observational model
\begin{equation}
    p(\bm y\mid\bm x)=\mN(\bm y;\mA(\bm x),\bm R)
\end{equation}
as an example. The DPS, DMPS, and SDA estimators for the term $p(\bm y\mid\bm x_t)$ are summarized in \cref{tab:conditional-likelihood}, where we use $\bm A$ to denote the linearization of the observational operator $\mA$.
\begin{table}
    \centering
    \caption{Comparison of various existing estimators for the adversarial gradient.}
    \label{tab:conditional-likelihood}
    \begin{tabular}{l|c|c}
        \toprule
        Method                           & $p_{0\mid t}(\bm x_0\mid\bm x_t)$                                           & $p_t(\bm y\mid\bm x_t)$                                                                           \\
        \midrule
        DPS \cite{Chung2023DPS}                   & $\delta(\bm x_0-\hat{\bm x}_0(\bm x_t,t))$                                  & $\mN(\bm y;\mA(\hat{\bm x}_0(\bm x_t,t)),\bm R)$                                                  \\
        \midrule
        DMPS \cite{Meng2022DMPS}                  & $\mN\left(\bm x_0;\mu_t^{-1}\bm x_t,\frac{\sigma_t^2}{\mu_t^2}\bm I\right)$ & $\mN\left(\bm y;\mu_t^{-1}\mA(\bm x_t),\bm R+\frac{\sigma_t^2}{\mu_t^2}\bm A\bm A^\trans\right)$  \\
        \midrule
        SDA \cite{Rozet2023sda,Rozet2023sda-2lqg} & N/A                                                                         & $\mN\left(\bm y;\mA(\hat{\bm x}_0(\bm x_t,t)),\bm R+\frac{\sigma_t^2}{\mu_t^2}\gamma\bm I\right)$ \\
        \bottomrule
    \end{tabular}
\end{table}
\subsection{State-observation augmented diffusion model}
\begin{figure}
    \centering
    \includegraphics[width=\textwidth]{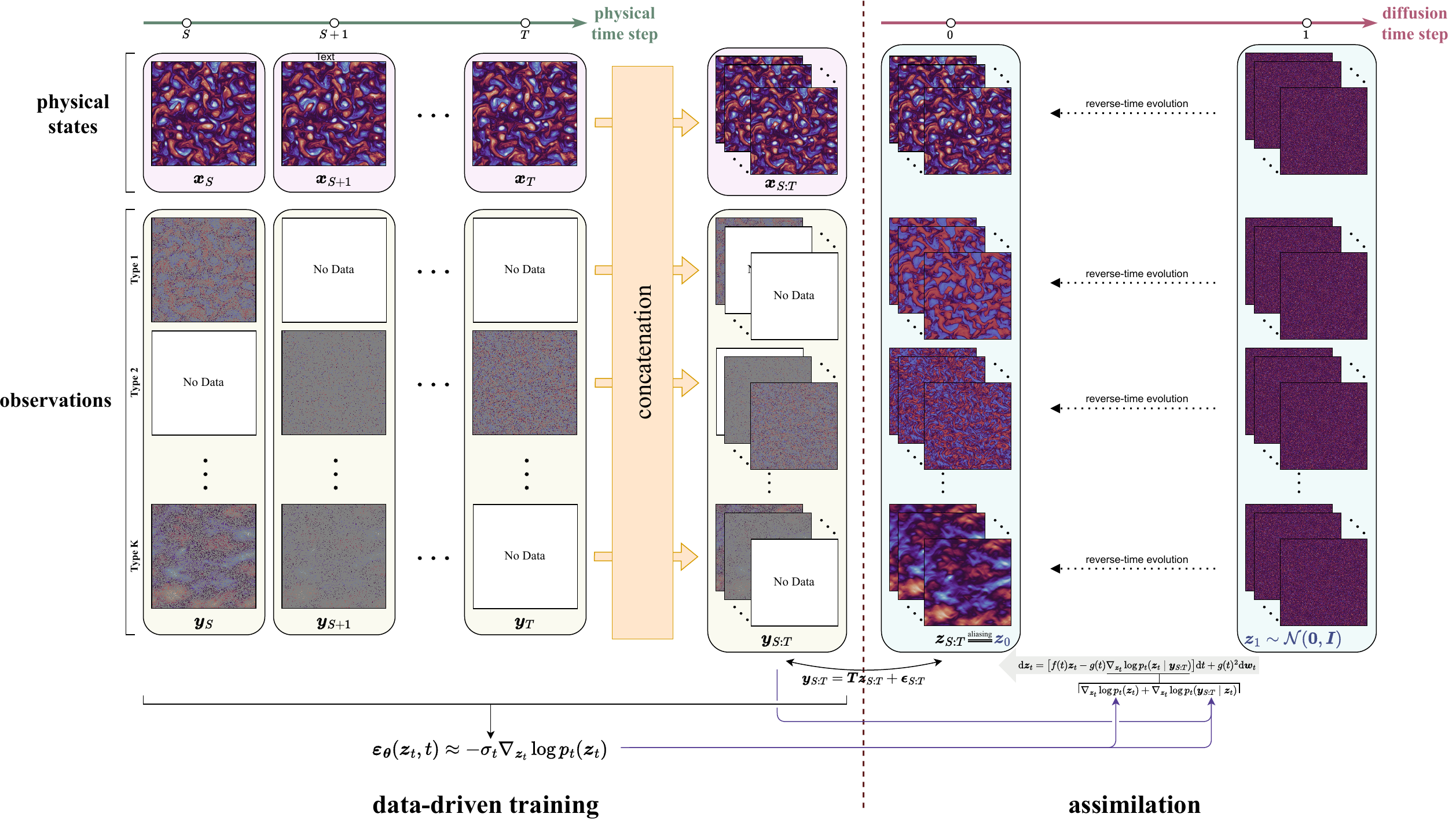}
    \caption{Illustration of assimilation with our State-Observation Augmented Diffusion model.}
    \label{fig:SOAD}
\end{figure}

In this section, we introduce our State-Observation Augmented Diffusion (SOAD) model, which is designed to handle the assimilation problem with nonlinear physical and observational models. We first present the state-observation augmented dynamical system as an equivalent but linear form of the original assimilation problem to alleviate the nonlinearity issue. Then, we derive an estimator for the marginal posterior $p_t(\bm x_t\mid\bm y)$ defined in \cref{eq:marginal-posterior}, which has been shown to match the real posterior under mild assumptions.
\Cref{fig:SOAD} depicts the general idea of our SOAD model, where we aim to assimilate on a time window from $S$ to $T$ with $K$ types of observations.
Finally, we introduce a forward-diffusion corrector to stabilize the reverse-time generation process of the diffusion model, followed by an overall implementation of the SOAD model.

\subsubsection{Augmented dynamical system}
To handle the potential nonlinearity of the observational model, we propose to consider an augmented version of \cref{eq:discrete_dynamical_system}. Let $\bm z_k$ denote the concatenation of the state variables $\bm x_k$ and all possible observation variables $\mH(\bm x_k)$, then we have
\begin{equation}\label{eq:augmented-model}
    \bm z_k=
    \begin{pmatrix}
        \bm x_k \\
        \mH(\bm x_k)
    \end{pmatrix}
    =
    \begin{pmatrix}
        \mM_k^X(\bm x_{k-1},\bm\eta_k) \\
        \mH\left(\mM_k^X(\bm x_{k-1},\bm\eta_k)\right)
    \end{pmatrix}
    =:\mM_k^Z\left(\bm z_{k-1},\bm\eta_k\right),\quad\bm\eta_k\sim\mathscr{D}^X,
\end{equation}
where the superscript ``$Z$'' emphasizes the augmented space where $\bm z_k$ lies.
By the assumptions in \cref{sec:problem-settings-assumptions}, for the observations at the $k$-th time step, there exists a linear mapping $\bm T_k$ linked to $\mH_k$ such that
\begin{equation}\label{eq:augmented-observation}
    \bm y_k=\mH_k(\bm x_k)+{\bm\epsilon_k}=\bm S_k\mH(\bm x_k)+{\bm\epsilon_k}=\bm T_k\bm z_k+{\bm\epsilon_k},\quad{\bm\epsilon_k}\sim\mathscr{D}^Y.
\end{equation}
Therefore, we have essentially established an equivalent augmented system
\begin{equation}\label{eq:augmented-dynamical-system}
    \begin{cases}
        \bm z_k=\mM_k^Z\left(\bm z_{k-1},\bm\eta_k\right), & \bm\eta_k\sim\mathscr{D}^X,        \\
        \bm y_k=\bm T_k\bm z_k+{\bm\epsilon_k},           & {\bm\epsilon_k}\sim\mathscr{D}^Y.
    \end{cases}
\end{equation}
Clearly, the forward propagation operator $\mM_k^Z$ inherits the periodicity of $\mM_k^X$. The observational operator $\bm T_k$ acting on $\bm z_k$ is now a time-dependent subsampling matrix, which simplifies the assimilation process.
By concatenating $\bm z$ and $\bm y$ over consecutive time steps as $\bm z_{S:T}$ and $\bm y_{S:T}$, respectively, the assimilation problem we focus on becomes a posterior estimation for $p\left(\bm z_{S:T}\mid\bm y_{S:T}\right)$. The prior $p\left(z_{S:T}\right)$ implicitly encodes the physical dynamics as well as the model error, and the observational model
\begin{equation}
    \bm y_{S:T}=\bm T_{S:T}\bm z_{S:T}+{\bm\epsilon_{S:T}},\quad{\bm\epsilon_{S:T}}\sim\mathscr{D}_{S:T}^Y
\end{equation}
becomes linear and Gaussian,
where we use the subscripts ``$S:T$'' to denote the concatenation of the corresponding variables, mappings, and distributions over the time interval from $S$ to $T$.

\subsubsection{Estimation for the marginal posterior}
In general, our State-Observation Augmented Diffusion (SOAD) model builds on the principles of score-based generative models outlined in \cref{sec:score-based-generative-models}.
Specifically, we construct a score-based generative model for $\bm z_{S:T}$ instead of $\bm x$.
For simplicity, we omit the subscript ``$S:T$'' hereafter, and thus $\bm z_{S:T}$ becomes $\bm z=\bm z_0$ in the context of diffusion process. We need to stress that the following derivations are all applied to concatenated versions of states, observations, and noise (distributions) over the time interval from $S$ to $T$.

To estimate the marginal posterior $p_t(\bm z_t\mid\bm y)$, we continue from the Bayes' rule
\begin{equation}\label{eq:conditional-score-bayes-augmented}
    \nabla_{\bm z_t}\log p_t(\bm z_t\mid\bm y)=\nabla_{\bm z_t}\log p_t(\bm z_t)+\nabla_{\bm z_t}\log p_t(\bm y\mid\bm z_t)
\end{equation}
adapted from \cref{eq:conditional-score-bayes,eq:conditional-likelihood} in the current context, where we still use the notation
\begin{equation}\label{eq:conditional-likelihood-augmented}
    \nabla_{\bm z_t}\log p_t(\bm y\mid\bm z_t)=\nabla_{\bm z_t}\log\int p(\bm y\mid\bm z)p_{0\mid t}(\bm z\mid\bm z_t)\md\bm z.
\end{equation}

On one hand, our SOAD model directly trains a neural network
\begin{equation}\label{eq:denoising-network}
    \bm\varepsilon_{\bm\theta}(\bm z_t,t)=-\sigma_t\bm s_{\bm\theta}(\bm z_t,t)\approx-\sigma_t\nabla_{\bm z_t}\log p_t(\bm z_t)
\end{equation}
according to \cref{eq:denoising-loss} to estimate the joint unconditional distribution for $\bm z$.

On the other hand, to estimate the adversarial gradient $\nabla_{\bm z_t}\log p_t(\bm y\mid\bm z_t)$, by applying Bayes' rule again, we have
\begin{equation}\label{eq:reverse-time-transition-bayes}
    \begin{aligned}
        \nabla_{\bm z}\log p_{0\mid t}(\bm z\mid\bm z_t) & =\nabla_{\bm z}\log p_{t\mid0}(\bm z_t\mid\bm z)+\nabla_{\bm z}\log p(\bm z)                \\
                                                         & =\sigma_t^{-2}\nabla_{\bm z}\left\|\bm z_t-\mu_t\bm z\right\|^2+\nabla_{\bm z}\log p(\bm z) \\
                                                         & =\mu_t^2\sigma_t^{-2}\left(\bm z-\mu_t^{-1}\bm z_t\right)+\nabla_{\bm z}\log p(\bm z).
    \end{aligned}
\end{equation}
If we make Gaussian assumptions for the prior $p(\bm z)$ with a covariance matrix $\bm\Sigma_0$, then $p_{0\mid t}(\cdot\mid\bm z_t)$ is also Gaussian by \cref{eq:reverse-time-transition-bayes}, whose covariance matrix has a closed form
\begin{equation}\label{eq:Sigma-0-mid-t}
    \bm\Sigma_{0\mid t}=\left(\bm\Sigma_0^{-1}+(\sigma_t^2/\mu_t^2)^{-1}\bm I\right)^{-1}.
\end{equation}
Similar to \cite{Chung2023DPS}, the mean of $p_{0\mid t}(\cdot\mid\bm z_t)$ may be provided by Tweedie's formula as
\begin{equation}
    \hat{\bm z}_0(\bm z_t,t)=\mE[\bm z_0\mid\bm z_t]=\mu_t^{-1}\left(\bm z_t+\sigma_t^2\nabla_{\bm z_t}\log p_t(\bm z_t)\right).
\end{equation}
Therefore, we use the following estimation for the reverse-time transition kernel
\begin{equation}
    \begin{aligned}
        p_{0\mid t}^{\mathrm{SOAD}}(\bm z\mid\bm z_t) & =\mN\left(\bm z;\hat{\bm z}_0(\bm z_t,t),\bm\Sigma_{0\mid t}\right)                                                                                               \\
                                                      & =\mN\left(\bm z;\mu_t^{-1}\left(\bm z_t-\sigma_t\bm\varepsilon_\theta(\bm z_t,t)\right),\left(\bm\Sigma_0^{-1}+(\sigma_t^2/\mu_t^2)^{-1}\bm I\right)^{-1}\right).
    \end{aligned}
\end{equation}
It follows that the term $p_t(\bm y\mid\bm z_t)$ can be computed analytically once the observational likelihood $p(\bm y\mid\bm z_0)$ is provided explicitly by \cref{eq:conditional-likelihood-augmented}. For instance, when the observational noise is Gaussian with covariance matrix $\bm\Sigma^Y$, we may assume
\begin{equation}\label{eq:Gaussian-observational-model}
    p(\bm y\mid\bm z)=\mN\left(\bm y;\bm T\bm z,\bm\Sigma^Y\right),
\end{equation}
and then the time-dependent marginal likelihood becomes
\begin{equation}\label{eq:SOAD-estimator}
    p^{\mathrm{SOAD}}(\bm y\mid\bm z_t)=\mN\left(\bm y;\bm T\hat{\bm z}_0(\bm z_t,t),\bm\Sigma^Y+\bm T\bm\Sigma_{0\mid t}\bm T^\trans\right).
\end{equation}
Note that despite the possible nonlinearity of the observational operator $\mH$, the augmentation as described in \cref{eq:augmented-dynamical-system} allows us to handle the assimilation problem with a linear observational model.

To sum up, we have the following theorem.
\begin{Thm}\label{thm:main}
    Under Gaussian assumption for the prior $p(\bm z)$ with covariance $\bm\Sigma_0$, our estimator given by \cref{eq:SOAD-estimator} \underline{matches} the marginal posterior $p_t(\bm y\mid\bm z_t)$ defined in \cref{eq:conditional-likelihood} when the observation noise $\mathscr D^Y$ is Gaussian in \cref{eq:augmented-observation}, or equivalently, the assumption in \cref{eq:Gaussian-observational-model} holds. Furthermore, for any scalar prior covariance $\bm\Sigma_0=\sigma_z^2\bm I$, our estimator becomes
    \begin{equation}\label{eq:SOAD-estimator-scalar}
        p_{\mathrm{SOAD}}(\bm y\mid\bm z_t)=\mN\left(\bm y;\bm T\hat{\bm z}_0(\bm z_t,t),\bm\Sigma^Y+\frac{\sigma_z^2r_t^2}{\sigma_z^2+r_t^2}\bm I\right),
    \end{equation}
    where we define $r_t=\sigma_t/\mu_t$.
\end{Thm}
\begin{proof}
    The first part is evident from the previous deduction. To show the second part, we may notice that $\bm\Sigma_{0\mid t}$ is scalar as long as $\bm\Sigma_0$ is scalar according to \cref{eq:Sigma-0-mid-t}, so it follows that
    \begin{equation}\label{eq:SOAD-scalar-covariance}
        \bm T{\bm\Sigma}_{0\mid t}\bm T^\trans=\frac{\sigma_z^2r_t^2}{\sigma_z^2+r_t^2}\bm T\bm T^\trans=\frac{\sigma_z^2r_t^2}{\sigma_z^2+r_t^2}\bm I,
    \end{equation}
    where we use the fact that $\bm T$ is a subsampling matrix and thus has orthonormal rows.
\end{proof}
\begin{Rem}
    We need to clarify that one should not confuse our SOAD estimator with the SDA estimator listed in \cref{tab:conditional-likelihood}. First, we only apply our estimator to the augmented model \eqref{eq:augmented-dynamical-system}, where the operator $\bm T$ is guaranteed to be linear. Under mild assumptions in \cref{thm:main}, the linearity ensures that the time-dependent marginal likelihood $p_t(\bm y\mid\bm z_t)$ remains Gaussian. In contrast, the SDA estimator employs a Gaussian approximation of the true $p_t(\bm y\mid\bm z_t)$, and the linearization of $\mH$ may introduce significant errors if the observational operator $\mH${ }is highly nonlinear. Furthermore, the approximation term $r_t^2\gamma\bm I$ in SDA does not prevent the covariance from exploding as the diffusion step $t$ approaches 1, resulting in a prior for $p(\bm y)$ with an unbounded covariance.
\end{Rem}
\subsubsection{Forward-diffusion corrector}
In our augmented model \eqref{eq:augmented-dynamical-system}, the observation $\bm y$ is a linear transformation of $\bm z$, or more specifically, a subsampled version of $\bm z$ with observational noise as described in \cref{eq:augmented-observation}. To stabilize the reverse-time diffusion process, we propose directly replacing part of the denoised results with a perturbed version of $\bm{y}$.
{The correction is based on an auxiliary forward diffusion of the observational data, which is then used to guide or adjust the denoising states where observations are available. Consequently, we refer to such a corrector as the ``forward-diffusion corrector''.}

More precisely, let $\bm z_t$ be the diffusion state at time $t$ with an initial state $\bm z_0=\bm z$, then
\begin{equation}
    \bm T\bm z_t=\bm T(\mu_t\bm z_0+\sigma_t\bm\varepsilon)\sim\mN\left(\mu_t\bm T\bm z,\sigma_t^2\bm I\right)
\end{equation}
with a standard Gaussian noise $\bm\varepsilon$ according to \cref{eq:transition-kernel}.
To calibrate the diffusion state $\bm z_t$, we propose substituting $\bm T\bm z_t$ at time $t$ with
$(\mu_t\bm y+\bm\varepsilon_t')$
assuming that \cref{eq:Gaussian-observational-model} holds,
where $\bm\varepsilon_t'\sim\mathscr{D}_t'$ is a random noise term. Since \cref{eq:Gaussian-observational-model} implies
\begin{equation}
    \mu_t\bm y\sim\mN\left(\mu_t\bm T\bm z,\mu_t^2\bm\Sigma^Y\right),
\end{equation}
we ensure that $\bm T\bm z_t$ and $(\mu_t\bm y+\bm\varepsilon_t')$ share the same means and covariances by setting
\begin{equation}
    \mathscr{D}_t'=\mN\left(\bm0,\sigma_t^2\bm I-\mu_t^2\bm\Sigma^Y\right)
\end{equation}
whenever {the matrix $\left(\sigma_t^2\bm I-\mu_t^2\bm\Sigma^Y\right)$} is positive semidefinite.

\begin{figure}
    \centering
    \includegraphics[width=\textwidth]{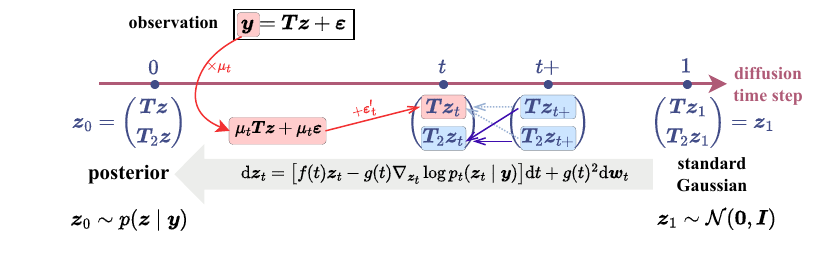}
    \caption{Illustration of the forward-diffusion corrector. }
    \label{fig:forward-diffusion-corrector}
\end{figure}

It is important to note that this condition generally holds except for small $t$, particularly when the observation noise is not excessively large. This follows from the fact that, under typical diffusion schedulers, $\lim_{t\to1^-}r_t=+\infty$ ensuring that our correction provides reliable estimation of $\bm T\bm z_t$ most of the time. The overall procedure is illustrated in \cref{fig:forward-diffusion-corrector}, where we define $\bm T_2$ as the compensated\footnote{The term ``compensated'' refers to the following construction. Given a high-dimensional row vector $\bm{x} \in \mathbb{R}^n$ and a subsampling matrix $\bm{T}$ acting on $\mathbb{R}^n$, we can always decompose $\bm{x}$ as $\bm x=(\bm{T} \bm{x}, \bm{x}_2)$, up to a permutation of the coordinate axes. The mapping $\bm{x} \mapsto \bm{x}_2$ is then called the compensated version of $\bm{T}$, representing the complement of the subsampling process.}
subsampling matrix associated with $\bm T$.

\subsubsection{Implementation}
We describe the detailed denoising procedure in this subsection, where we set ${\mathscr D}^Y=\mN(\bm0,(\sigma^\mathrm{o})^2\bm I)$ for simplicity. Similar deduction can be applied to other more complex settings.

Let $r_\sigma=\sigma_{t_-}/\sigma_t$ and $r_\mu=\mu_{t_-}/\mu_t$ be the ratios of the diffusion parameters at adjoint time step $t_-$ and $t$, respectively. We choose the EI discretization scheme
\begin{equation}
    \bm z_{t_-}=r_\mu\bm z_t+\left(r_\sigma-r_\mu\right)\bm\pi_\theta(\bm z_t,t)
\end{equation}
as in \cref{eq:EI-predictor} for the reverse-time stepper, where we define $\bm\pi_\theta(\bm z_t,t)=-\sigma_t^2\nabla_{\bm z_t}\log p_t(\bm z_t\mid\bm y)$.
By plugging \cref{eq:conditional-score-bayes-augmented,eq:denoising-network,eq:SOAD-estimator-scalar} into the definition of $\bm\pi_\theta$, we have
\begin{equation}\label{eq:pi-theta}
    \begin{aligned}
        \bm\pi_\theta(\bm z_t,t) & =-\sigma_t^2\nabla_{\bm z_t}\log p_t(\bm z_t\mid\bm y)                                                                                                                                                                                 \\
                                 & =-\sigma_t^2\nabla_{\bm z_t}\log p_t(\bm z_t)-\sigma_t^2\nabla_{\bm z_t}\log p_t(\bm y\mid\bm z_t)                                                                                                                                     \\
                                 & =\sigma_t\bm\varepsilon_\theta(\bm z_t,t)+\frac{\sigma_t^2/2}{(\sigma^\mathrm{o})^2+\frac{\sigma_z^2r_t^2}{\sigma_z^2+r_t^2}}\nabla_{\bm z_t}\left\|\bm y-\bm T\hat{\bm z}_0(\bm z_t,t)\right\|^2                                      \\
                                 & =\sigma_t\bm\varepsilon_\theta(\bm z_t,t)+\frac{r_t^2/2}{(\sigma^\mathrm{o})^2+\frac{\sigma_z^2r_t^2}{\sigma_z^2+r_t^2}}\nabla_{\bm z_t}\left\|\mu_t\bm y-\bm T\left(\bm z_t-\sigma_t\bm\varepsilon_\theta(\bm z_t,t)\right)\right\|^2 \\
                                 & =\sigma_t\bm\varepsilon_\theta(\bm z_t,t)+c_t\nabla_{\bm z_t}\left\|\mu_t\bm y-\bm T\bm z_t+\bm T\sigma_t\bm\varepsilon_\theta(\bm z_t,t)\right\|^2
    \end{aligned}
\end{equation}
In practice, we treat $\sigma_z$ as a hyperparameter to be tuned as it is hard to evaluate explicitly for a trained generative model. Meanwhile, a certain clipping mechanism for the coefficient $c_t$ is required to stabilize the generation process when the diffusion step $t$ is near 1. Fortunately, the model performances are not sensitive to the choice of $\sigma_z$ as well as the clipping mechanism. See \cref{alg:soad-denoising} for the complete implementation.

\begin{algorithm}
    \setstretch{1.2}
    \centering
    \caption{Assimilation with the SOAD model for Gaussian noise}
    \label{alg:soad-denoising}
    \footnotesize
    \begin{algorithmic}[1]
        \State {\textbf{Input:} The subsampling matrix $\bm T$, the observation data $\bm y$, the noise distribution ${\mathscr D}^Y=\mN\left(\bm0,(\sigma^\mathrm{o})^2\bm I\right)$, a trained noise estimator $\bm\varepsilon_\theta(\bm z_t,t)$, and the denoising diffusion time step $\delta t$;}
        \State \textbf{Hyperparameters:} $\sigma_z=1.0$, $\delta=0.25$, $N_c=5$.
        \vspace*{2ex}
        \Function{AssimilationWithSOAD}{$\bm T$, $\bm y$, $\sigma^\mathrm{o}$, $\bm\varepsilon_{\bm\theta}$, $\delta t$}
        \State $t\leftarrow1$; $\bm z_t\sim\mN(0,\bm I)$;
        \While{$t>0$}
        \vspace*{2ex}
        \State $t_-\leftarrow t-\delta t$; $r_\sigma\leftarrow\sigma_{t_-}/\sigma_t$; $r_\mu\leftarrow\mu_{t_-}/\mu_t$;
        \State $c_t\gets\frac{r_t^2/2}{(\sigma^\mathrm{o})^2+\frac{\sigma_z^2r_t^2}{\sigma_z^2+r_t^2}}$;
        \State $\bm Q_t\gets\nabla_{\bm z_t}\left\|\mu_t\bm y-\bm T\bm z_t+\bm T\sigma_t\bm\varepsilon_\theta(\bm z_t,t)\right\|^2$;
        \State $\hat c_t\gets\max\left(1,1/\|\bm Q_t\|_\infty\right)$;\Comment{clipping mechanism}
        \State $\bm\pi_\theta(\bm z_t,t)\gets\sigma_t\bm\varepsilon_\theta(\bm z_t,t)+\hat c_t\bm Q_t$;\Comment{by \cref{eq:pi-theta}}
        \State $\bm z_{t_-}\gets r_\mu\bm z_t+\left(r_\sigma-r_\mu\right)\bm\pi_\theta(\bm z_t,t)$;\Comment{reverse-time evolution}
        \State $\bm z_{t_-}\leftarrow$ \Call{FDC}{$\bm z_{t_-}$, $t_-$};
        \vspace*{2ex}
        % \State /// Apply the corrector by making a few steps of LMC sampling as in \cref{eq:LMC-corrector}.
        \State $\bm s_{t_-}\gets\sigma_{t_-}^{-1}\bm\pi_\theta(\bm z_{t_-},t)$;\Comment{cache the conditional score function}
        \For{$i=1$ to $N_c$}
        \State $\bm z_{t_-}\gets\bm z_{t_-}+\frac{\delta}{2}\bm s_{t_-}+\sqrt\delta\bm\varepsilon'',\quad\bm\varepsilon''\sim\mN(0,\bm I)$;\Comment{LMC sampling as in \cref{eq:LMC-corrector}}
        \State $\bm z_{t_-}\gets$ \Call{FDC}{$\bm z_{t_-}$, $t_-$};
        \EndFor
        \vspace*{2ex}
        \State $t\gets t_-$;
        \EndWhile
        \EndFunction
        \vspace*{2ex}
        \Function{FDC}{$\bm z_t$, $t$}\Comment{the forward-diffusion corrector}
        \State $r_t\leftarrow\sigma_t/\mu_t$;
        \If{$\sigma^\mathrm{o}\le r_t$}
        \State $\bm T\bm z_t\gets\mu_t\left(\bm y+\sqrt{r_t^2-(\sigma^\mathrm{o})^2}\bm\varepsilon_t'\right),\quad \bm\varepsilon_t'\sim\mN(0,\bm I)$;
        \EndIf
        \State \textbf{return} $\bm z_t$
        \EndFunction
    \end{algorithmic}
\end{algorithm}

% section 4

\section{Experiments}\label{sec:experiments}
To show the advantages of our proposed SOAD model, we conduct experiments on the two-layer quasi-geostrophic model following the settings proposed in the Score-based Data Assimilation (SDA) \cite{Rozet2023sda,Rozet2023sda-2lqg},
and we treat it as the baseline due to its effectiveness shown for linear observations.

We start by outlining the dataset and the network training procedures for the experiments. Next, we detail the observation operators as well as the assimilation settings we used for testing. A comparative analysis of our SOAD model and the SDA baseline with various observational operators is presented. Besides, we also try to explore the long-term behavior of our model and its capability of handling multiple observations simultaneously. Our experimental results indicate that our SOAD model is more suitable for dealing with highly nonlinear observations.

\subsection{Dataset}
Consider the two-layer quasi-geostrophic (QG) evolution equation
\begin{equation}\label{eq:qg-model}
    \begin{aligned}
        \partial_tq_1+J(\psi_1,q_1)+\beta_1\psi_{1x} & =\mathrm{ssd},                      \\
        \partial_tq_2+J(\psi_2,q_2)+\beta_2\psi_{2x} & =-r_{ek}\nabla^2\psi_2+\mathrm{ssd}
    \end{aligned}
\end{equation}
with the potential vorticity
\begin{equation}\label{eq:q-p-laplacian}
    q_1=\nabla^2\psi_1+F_1(\psi_2-\psi_1),\quad q_2=\nabla^2\psi_2+F_2(\psi_1-\psi_2)
\end{equation}
and the mean potential vorticity gradients
\begin{equation}
    \beta_1=\beta+F_1(U_1-U_2),\quad\beta_2=\beta-F_2(U_1-U_2).
\end{equation}
The horizontal Jacobian is defined as
\begin{equation}
    J(\psi,q)=\psi_x q_y-\psi_y q_x,
\end{equation}
and ``$\mathrm{ssd}$'' indicates the small-scale dissipation. To accelerate the preparation process, the dataset has been generated by a Python package \texttt{pyqg-jax}\footnote{\url{https://github.com/karlotness/pyqg-jax}} as a \texttt{Jax} \cite{jax2018github} implementation of the original \texttt{pyqg} \cite{pyqg2022github} library, and all the other equation parameters follow the default configurations (See \cref{app:qg-config} for completeness).

To create our dataset, we set the solution domain as $512\times512$ with 15-minute time steps. The QG equation is evolved from 10 different random initial states separately. After a warm-up period of 5 years, the reference states are downsampled from the trajectories to a spatial resolution of $256\times256$ with a temporal resolution of $\Delta t=24$ hours for 100,000 model days, which leads to a dataset of size $10\times100000\times2\times256\times256$ in total.
Owing to the chaotic nature of the QG model, the 5-year warm-up stage ensures that the trajectories evolve into statistically independent states, making it reasonable to directly split them into training and testing datasets. This setup allows the performance on the testing dataset to reflect the model's ability to generalize to unseen states, thereby assessing its generalization ability.

\begin{figure}
    \centering
    \includegraphics[width=.6\textwidth]{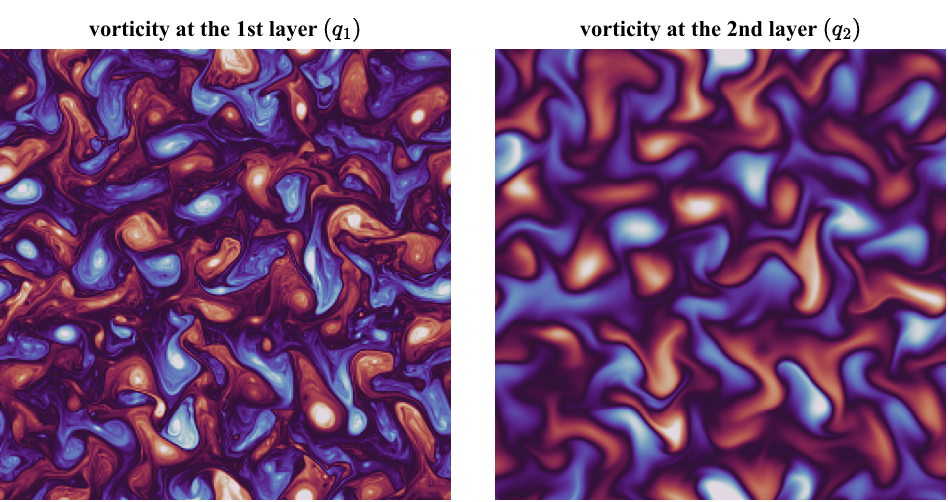}
    \caption{Visualization of the vorticity for the two-layer quasi-geostrophic model.}
    \label{fig:qg-demo}
\end{figure}

Since the evolution of the two-layer QG model does not depend on time, we assume that the vorticity snapshots $(q_1,q_2)^\trans$ of shape $(2, 256, 256)$ within a fixed time range follow the same data distribution. Consequently, we divide each trajectory into 1,000 windows and retain only the first 32 snapshots from every 100 snapshots to avoid temporal correlations. Such a process transforms the dataset into $10\times1000$ windows of size $32\times2\times256\times256$.  \cref{fig:qg-demo} visualizes the generated data for a single time step.
\subsection{Network structures and training procedures}
To ensure a fair comparison with the SDA baseline, we adopt the same network structure for the denoising network $\bm\varepsilon_\theta(\bm x_t,t)$, which utilizes a U-Net structure mixed with temporal convolutions and embeddings for diffusion time steps $t$. We have included the detailed architecture in \cref{sec:network-arch} for completeness. The only difference between our network and the SDA baseline lies in the input channels. SDA learns a generative model for the state $\bm x_{S:T}$, while our SOAD learns for the augmented state $\bm z_{S:T}${, which means that the corresponding observations $\bm y_{S:T}$ is fed into the network for training as well}. Since each augmented state includes a clean observation $\mH(\bm x)_{S:T}$, training SOAD requires additional calls of the observation operator $\mH$.

We use the first $8\times1000$ windows for training and reserve the remaining $2\times1000$ windows for validation and testing.
During training, a chunk of length 9 from each window $\bm z_{0:31}^{(j)}$ of length 32 is selected randomly to create a mini-batch for the networks, which gives rise to the following minimization problem
\begin{equation}\label{eq:denoising-loss-empirical}
    \min_{\bm\theta}L(\bm\theta)=\min_{\bm\theta}\frac{1}{2J}\sum_{j=1}^J\left\|\bm\varepsilon_{\bm\theta}\left(\mu_{t_j}\bm z_{S_j:S_j+8}^{(j)}+\sigma_{t_j}\bm\varepsilon_j,t_j\right)-\bm\varepsilon_j\right\|^2
\end{equation}
as an empirical estimation of \cref{eq:denoising-loss}, where
\begin{equation}
    \bm\varepsilon_j\sim\mN(\bm0,\bm I),\quad t_j\sim\mU_{[0,1]},\quad S_j\sim\mU_{\{0,1,\cdots,23\}}
\end{equation}
are all independent random variables for each $j$. Here, $J=8\times1000$ denotes the number of total training windows.
All networks are trained with the Adam optimizer \cite{Kingma2015Adam}, employing a learning rate of $2\times10^{-4}$ and a weight decay of $1\times10^{-5}$. To stabilize the training process, all the network inputs are normalized using the empirical mean and variance of the entire training dataset.

\subsection{Observations}
We consider the following Gaussian observational model
\begin{equation}\label{eq:observational-model}
    \bm y_k=\bm S_k\circ\mH(\bm x_k)+{\bm\epsilon_k},\quad{\bm\epsilon_k}\sim\mN\left(0,0.1^2\bm I\right),
\end{equation}
where $k$ denotes the time step index. $\bm x_k$ consists of the two-layer vorticity at time step $k$, and $\bm y_k$ stands for the associated noisy observations. There are three different observation operators used in our experiments to evaluate the performance of our SOAD model under various observational settings, namely,
\begin{itemize}
    \item an easy nonlinear ``arctan'' operator
          \begin{equation}
              \mH^\mathrm{e}:\bm x\mapsto\arctan3\bm x,
          \end{equation}
    \item a hard nonlinear sinusoidal operator
          \begin{equation}
              \mH^\mathrm{h}:\bm x\mapsto\frac32\sin3\bm x,\quad\text{and}
          \end{equation}
    \item a vorticity-to-velocity mapping
          \begin{equation}
              \mH^\mathsf{v2v}:\bm x=(q_1,q_2)^\trans\mapsto C\odot(u_1,v_1,u_2,v_2)^\trans.
          \end{equation}
\end{itemize}
The first two operators are element-wise, and we refer to $\mH^\mathrm{e}$ as the ``easy'' case since it is injective and monotonic, while $\mH^\mathrm{h}$ is termed the ``hard'' case due to its highly nonlinear behavior. Besides, to explore the potential of our model for real applications, we also consider the non-local vorticity-to-velocity mapping $\mH^\mathsf{v2v}$,
where $(u_i,v_i)$ are the velocity components for the $i$-th layer and $C\odot$ denotes a multiplication with component-wise scalars so that the outputs approximately have zero means and unit deviations. Such a choice is inspired by the fact that velocity components are often directly observed rather than the related vorticity in the real world.
To compute $\mH^\mathsf{v2v}$, \cref{eq:q-p-laplacian} is solved for the stream function $(\psi_1,\psi_2)^\trans$ in the spectral space, and then the velocity components are calculated by the stream functions. Note that we cannot conclude whether $\mH^\mathsf{v2v}$ is easier or harder, as the operator is linear, but the mapping rule is much more complex.

\begin{figure}
	\centering
	\includegraphics[width=\textwidth]{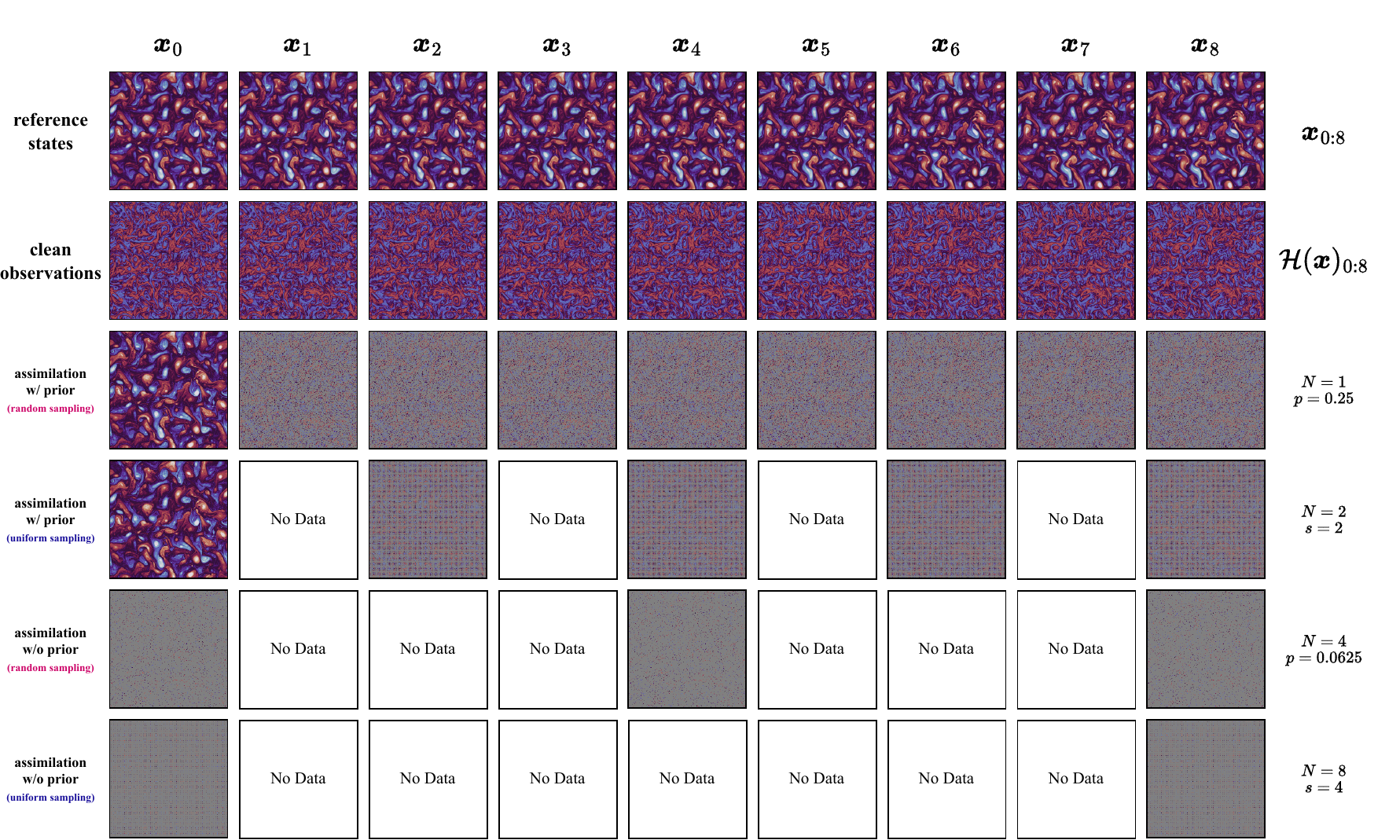}
	\caption{Illustration of different assimilation settings. (zoom-in for better visualization)}
	\label{fig:assimilation-and-forecast}
\end{figure}

Next, we need to specify the subsampling matrix $\bm S_k$ to determine the observational operator $\mH_k=\bm S_k\circ\mH$ for each $k$.
The observational frequency plays a crucial role, as different data sources, such as radar or satellite imagery, operate at varying time intervals. In our experiments, we denote by $N$ the number of time steps between consecutive observations and vary it across different settings.
We also investigate the impact of different observational ratios for partial observations, considering two masking strategies:
\begin{enumerate}[(i)]
    \item random sampling from the grid with a ratio $p$, and
    \item uniform sampling with a stride $s$, corresponding to an observational ratio\footnote{For a spatial stride of $s=10$, the actual observational ratio slightly exceeds 1\% since the snapshot size is not divisible by 10, but we omit this minor discrepancy for simplicity.} of $(1/s)^2$.
\end{enumerate}
Readers may refer to \cref{fig:assimilation-and-forecast} for detailed illustrations{, where we only visualize the first-layer vorticity ($q_1$) for simplicity}. Throughout our experiments, we evaluate random ratios $p\in\{1.0,0.25,0.625,0.01\}$ and strides $s\in\{1,2,4,10\}$ for partial observations under varying time intervals $N\in\{1,2,4,8\}$.

\subsection{Metric}
To make quantitative comparisons, we use the rooted mean-square error (RMSE)
\begin{equation}
    \mathrm{RMSE}(\bm x^\mathrm{r},\bm x^\mathrm{a})=\left(\frac{1}{2H_\mathrm{res}W_\mathrm{res}}\sum_{i=1}^2\sum_{j=1}^{H_\mathrm{res}}\sum_{k=1}^{W_\mathrm{res}}\left[\bar q_i^\mathrm{r}(j,k)-\bar q_i^\mathrm{a}(j,k)\right]^2\right)^{1/2}
\end{equation}
to evaluate the assimilated state $\bm x^\mathrm{a}$. Recall that $\bm x$ consists of vorticity $(q_1,q_2)^\trans$ on two levels, and here we use $(\bar q_1,\bar q_2)^\trans$ for their normalized versions. $\bm x^\mathrm{r}$ is the reference solution, and $(H_\mathrm{res},W_\mathrm{res})$ stands for the spatial resolution. Since all the data have been normalized in advance, the RMSEs result from different models and various assimilation settings are comparable. To decrease randomness, all the following experiments have been run for 5 times with different random seeds, and the averaged RMSEs are reported unless specified otherwise.
\subsection{Assimilation experiments}
In the assimilation experiments, we fix our assimilation window to 9 time steps, which means we have observations from at most 9 consecutive steps.
Our goal is to assimilate all state variables within this window, as illustrated in \cref{fig:assimilation-and-forecast}.

We have set the Score-based Data Assimilation (SDA) method as our baseline, which is a state-of-the-art method for data assimilation with data-driven physical models independent of any classical assimilation algorithms. Both the SOAD model and the SDA baseline are pre-trained with the same dataset and network structures, and no additional training is performed during the assimilation process. Besides, it is worth noting that knowledge of physical dynamics have been implicitly embedded into the SDA or our SOAD model through the training process, and only the observational models are used during the assimilation.

We consider the following {four} assimilation settings.
\begin{enumerate}[(I)]
    \item \textbf{Assimilation with a background prior} at the first step, which aligns with practical real-world applications.
    \item \textbf{Assimilation without any background prior}, where no background estimation is available at any step, and all physical prior knowledge is inferred from our trained SOAD model. This setting is more challenging and indicates the long-term performance of our model to some extent.
    \item \textbf{Multi-modal assimilation} as an extension of setting (II), where observations coming from multiple observational operators are available simultaneously.
    \item {\textbf{Assimilation with non-Gaussian distributions} as an extension of setting (I) and (II), where we set the background prior and the observational noise as non-Gaussian to study the robustness of our model.}
\end{enumerate}

% \subsubsection{Baseline}
% 
\subsubsection{Assimilation with background prior}\label{sec:ass-with-prior}
In practical applications, assimilation is usually performed sequentially once new observations become available, so we start with the classical assimilation settings that additionally provide the models with a background prior for the first step. We choose the background prior as an ideal Gaussian perturbation with a known covariance $(\sigma^\mathrm{b})^2=0.1^2$. Formally, the observational model for the first step is modified as
\begin{equation}\label{eq:background-prior}
    \bm y_0=\bm x_0+{\bm\epsilon_0},\quad{\bm\epsilon_0}\sim\mN\left(0,(\sigma^\mathrm{b})^2\bm I\right),
\end{equation}
where $\bm y_0$ is the background prior with noise variance $0.1^2$. See \cref{fig:assimilation-and-forecast} (3rd and 4th rows) for visualization.

\begin{figure}
    \centering
    \begin{subfigure}[t]{.33\textwidth}
        \centering
        \includegraphics[width=\textwidth]{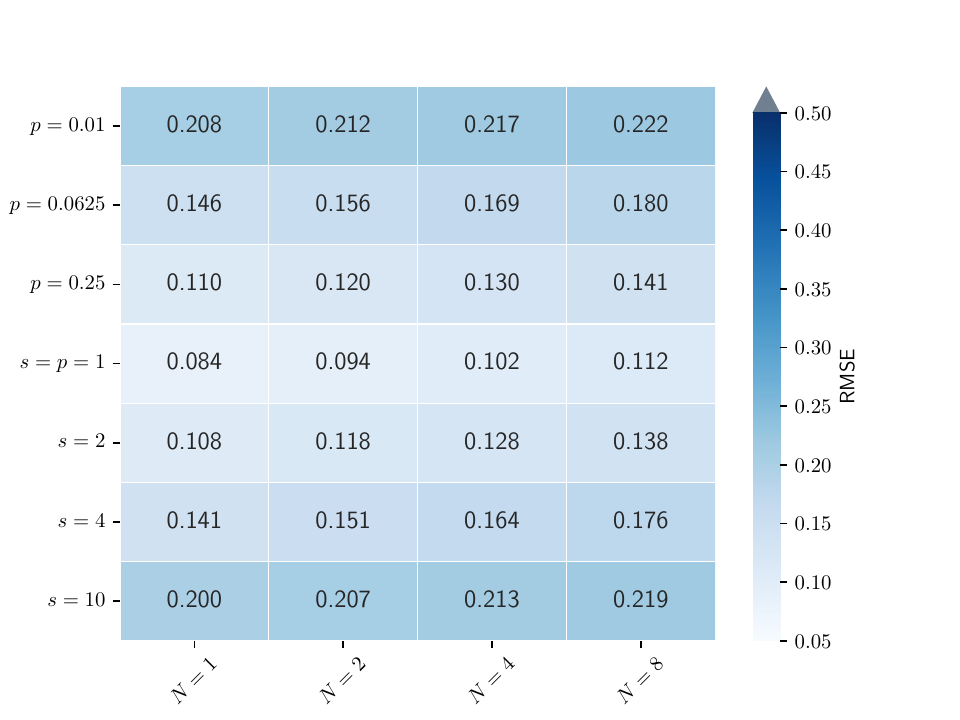}
        \caption{SOAD (ours) for $\mH^\mathrm{e}$}
        \label{fig:assimilate-with-prior-arctan3x-rmse-heatmap}
    \end{subfigure}%
    \begin{subfigure}[t]{.33\textwidth}
        \centering
        \includegraphics[width=\textwidth]{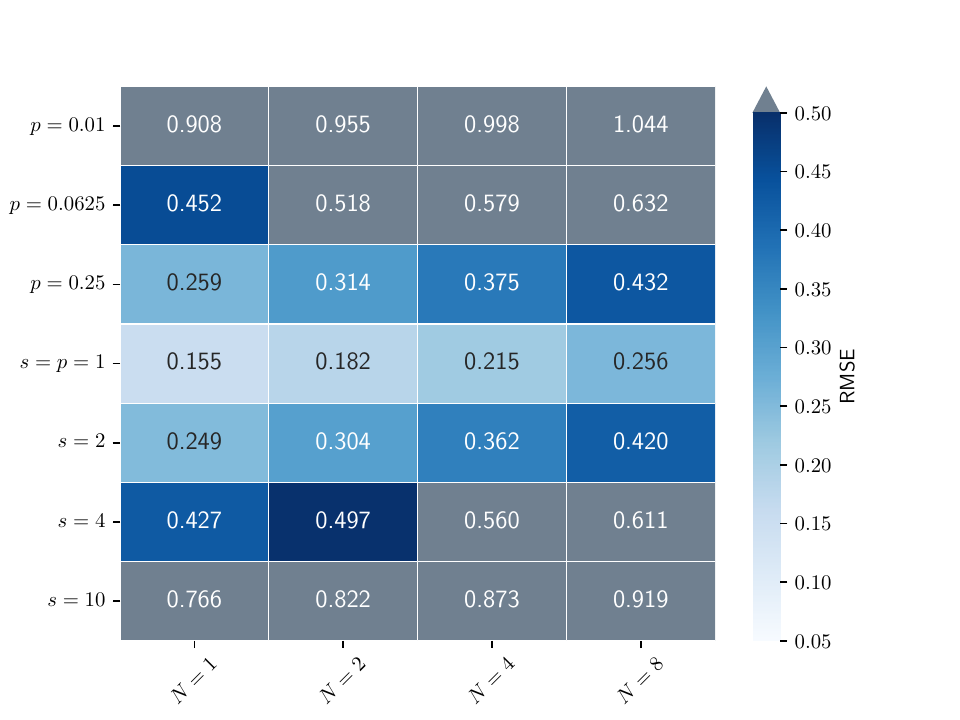}
        \caption{{SOAD (no f.d.c.) for $\mH^\mathrm{e}$}}
        \label{fig:assimilate-with-prior-arctan3x-no-fdc-rmse-heatmap}
    \end{subfigure}%
    \begin{subfigure}[t]{.33\textwidth}
        \centering
        \includegraphics[width=\textwidth]{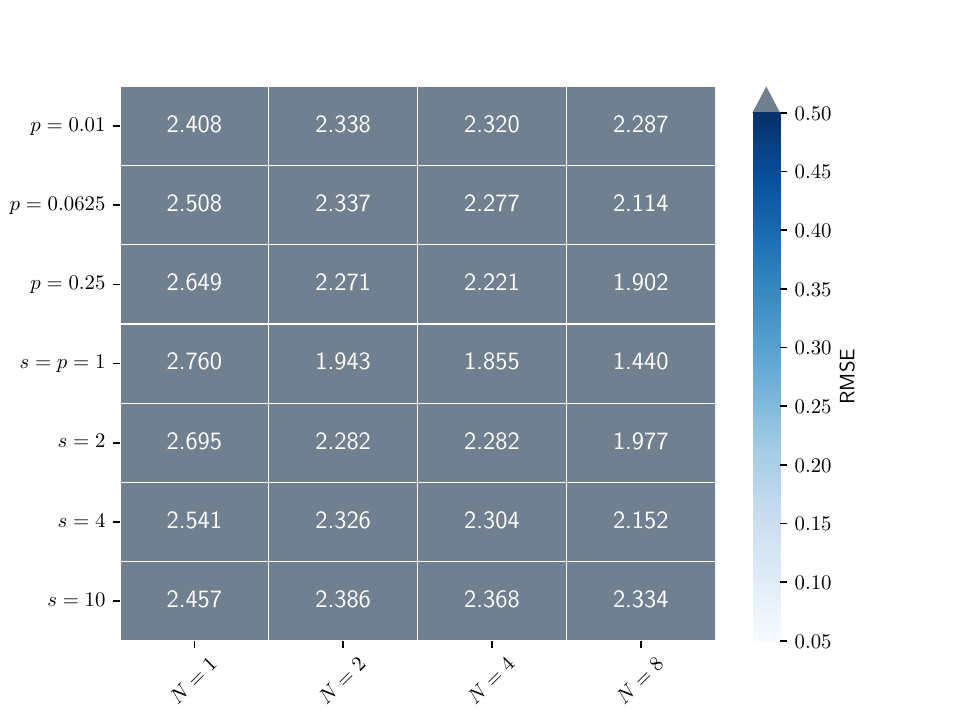}
        \caption{SDA (baseline) for $\mH^\mathrm{e}$}
        \label{fig:assimilate-with-prior-arctan3x-sda-heatmap}
    \end{subfigure}
    \begin{subfigure}[t]{.33\textwidth}
        \centering
        \includegraphics[width=\textwidth]{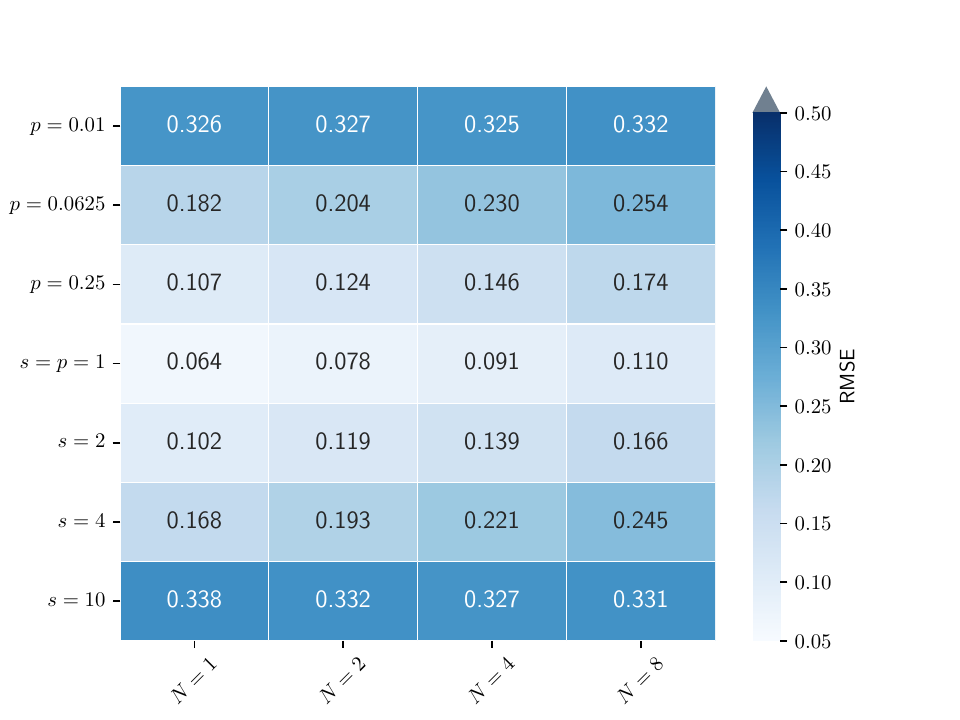}
        \caption{SOAD (ours) for $\mH^\mathrm{h}$}
        \label{fig:assimilate-with-prior-sin3x-rmse-heatmap}
    \end{subfigure}%
    \begin{subfigure}[t]{.33\textwidth}
        \centering
        \includegraphics[width=\textwidth]{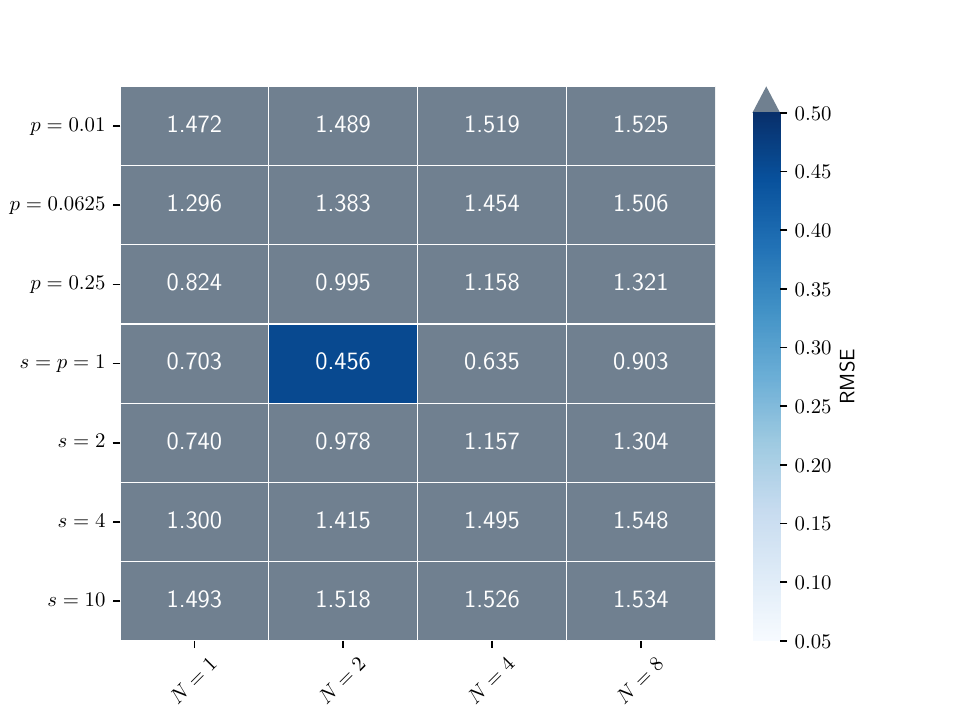}
        \caption{{SOAD (no f.d.c.) for $\mH^\mathrm{h}$}}
        \label{fig:assimilate-with-prior-sin3x-no-fdc-rmse-heatmap}
    \end{subfigure}%
    \begin{subfigure}[t]{.33\textwidth}
        \centering
        \includegraphics[width=\textwidth]{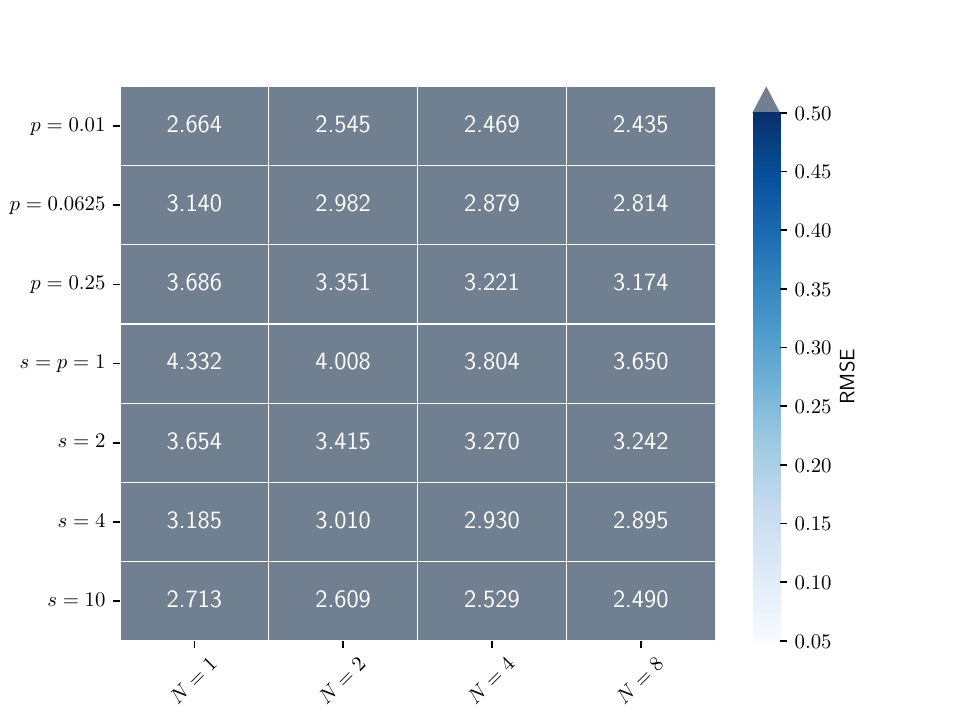}
        \caption{SDA (baseline) for $\mH^\mathrm{h}$}
        \label{fig:assimilate-with-prior-sin3x-sda-heatmap}
    \end{subfigure}
    \begin{subfigure}[t]{.33\textwidth}
        \centering
        \includegraphics[width=\textwidth]{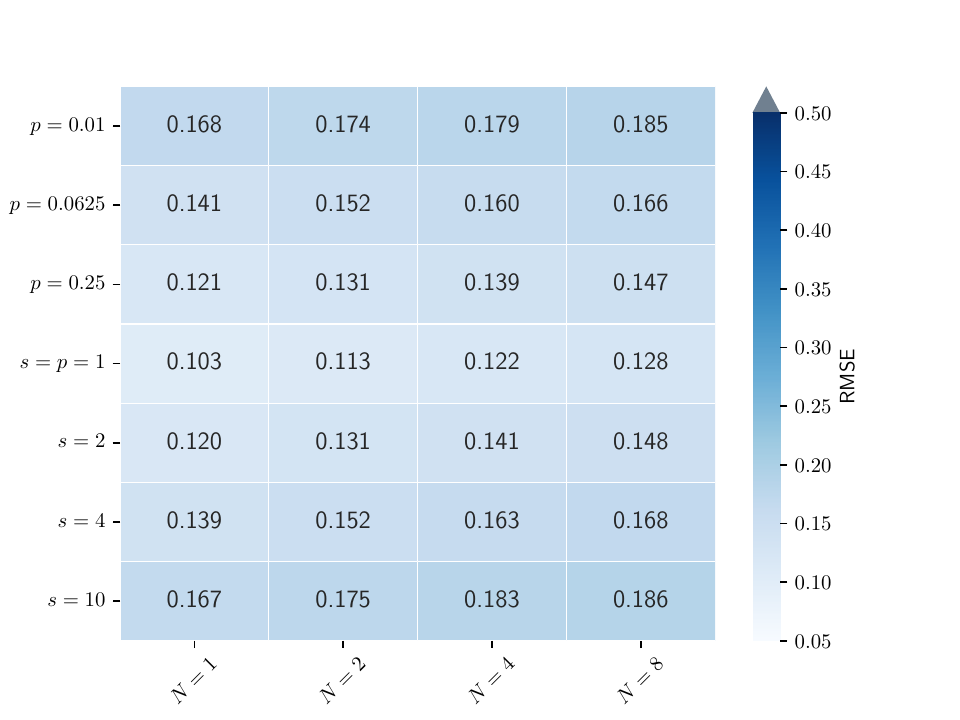}
        \caption{SOAD (ours) for $\mH^\mathsf{v2v}$}
        \label{fig:assimilate-with-prior-vor2vel-rmse-heatmap}
    \end{subfigure}%
    \begin{subfigure}[t]{.33\textwidth}
        \centering
        \includegraphics[width=\textwidth]{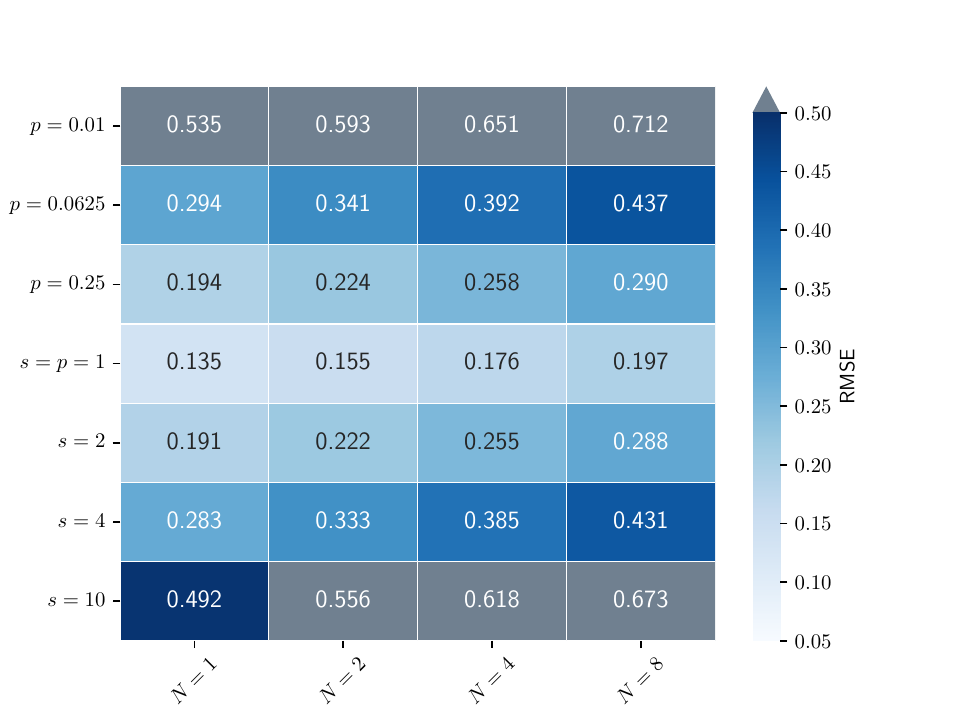}
        \caption{{SOAD (no f.d.c.) for $\mH^\mathsf{v2v}$}}
        \label{fig:assimilate-with-prior-vor2vel-no-fdc-rmse-heatmap}
    \end{subfigure}%
    \begin{subfigure}[t]{.33\textwidth}
        \centering
        \includegraphics[width=\textwidth]{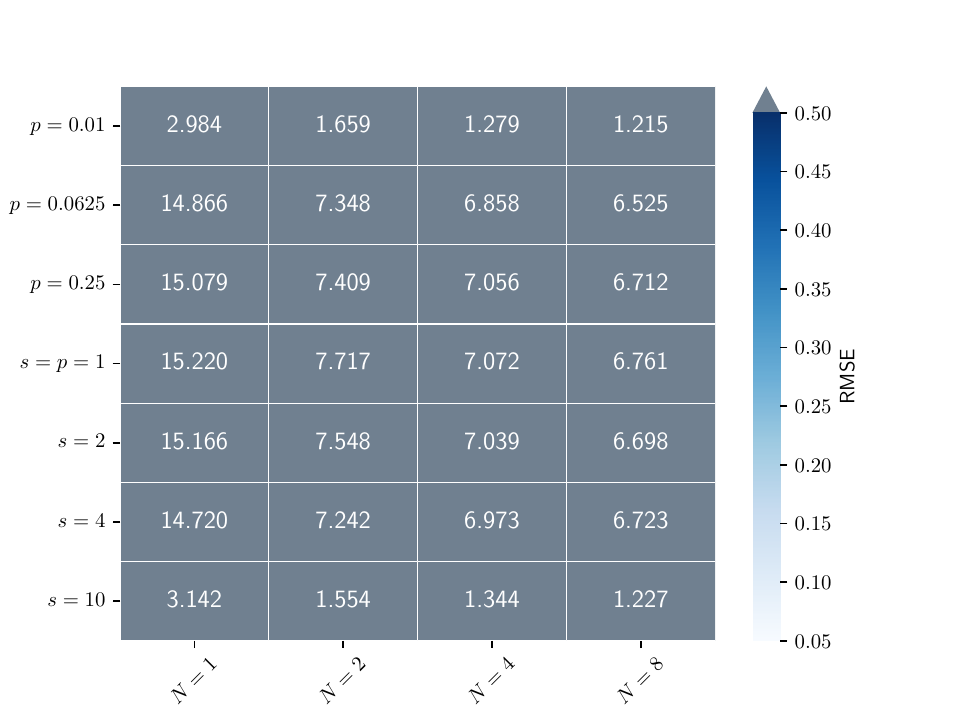}
        \caption{{SDA (baseline) for $\mH^\mathsf{v2v}$}}
        \label{fig:assimilate-with-prior-vor2vel-sda-heatmap}
    \end{subfigure}%
    \caption{Averaged RMSEs for the assimilated states by using background prior.}
    \label{fig:assimilate-with-prior-rmse-heatmap}
\end{figure}

Our experiments start with the element-wise observations $\mH^\mathrm{e}$ and $\mH^\mathrm{h}$.
{Besides, to explore the potential of our SOAD model in real applications, we also test it with the vorticity-to-velocity mapping $\mH^\mathsf{v2v}$.}
The averaged RMSEs over all the 9 time steps for the assimilated states are summarized in
\cref{fig:assimilate-with-prior-rmse-heatmap}.
Our SOAD model performs slightly worse as the time interval $N$ increases and the observational ratio $p$ or $(1/s)^2$ decreases, which aligns with our intuition that the assimilation task becomes more difficult with fewer observations. In the most challenging case, where the observations are available only every 8 time steps with a ratio of $1\%$, the RMSEs are around 0.2 and 0.3 for the easy and hard observations, respectively. This indicates that our method is still able to extract the physical features from rare observations to some extent. 
Another interesting observation is that the RMSEs for the hard observation $\mH^\mathrm{h}$ are lower than those for $\mH^\mathrm{e}$ and $\mH^\mathsf{v2v}$. One possible explanation is that when the observational data is sufficient for the model to capture the dynamics, observations from a harder operator $\mH^\mathrm{h}$ may provide more detailed information since $\mH^\mathrm{h}$ is much more sensitive to the input.
In contrast, the SDA baseline {(\mbox{\cref{fig:assimilate-with-prior-arctan3x-sda-heatmap,fig:assimilate-with-prior-sin3x-sda-heatmap,fig:assimilate-with-prior-vor2vel-sda-heatmap}})} shows a significant performance drop even when the observations are available everywhere for each time step. The inferior performance is likely due to the linearity assumption for the observation operator in the SDA formulations, which is unsuitable for highly nonlinear observations.

{
Meanwhile, \mbox{\cref{fig:assimilate-with-prior-rmse-heatmap}} includes an ablation study of our forward-diffusion corrector, abbreviated as ``f.d.c.'' in the captions, as well. To study the effectiveness of our forward-diffusion corrector, we have kept all the experimental settings the same as those for the SOAD and the SDA methods, but removed all the updates by the forward-diffusion corrector (line 12 and 16 in \mbox{\cref{alg:soad-denoising}}). By comparing \mbox{\cref{fig:assimilate-with-prior-arctan3x-rmse-heatmap,fig:assimilate-with-prior-sin3x-rmse-heatmap,fig:assimilate-with-prior-vor2vel-rmse-heatmap}} with \mbox{\cref{fig:assimilate-with-prior-arctan3x-no-fdc-rmse-heatmap,fig:assimilate-with-prior-sin3x-no-fdc-rmse-heatmap,fig:assimilate-with-prior-vor2vel-no-fdc-rmse-heatmap}} respectively we can conclude that the forward-diffusion corrector has played an important role in improving the assimilation performance. For the observations $\mH^\mathrm{e}$ and $\mH^\mathsf{v2v}$, our SOAD model without the forward-diffusion corrector has shown a similar pattern as the normal SOAD, which is to say, the denser and the more frequent observations we have for assimilation, the higher accuracy we can obtain. For the much harder observation $\mH^\mathrm{h}$, the forward-diffusion corrector has prevented divergence and thus exhibited its great potential to stablize the generative process.
}

\begin{figure}
    \centering
    \begin{subfigure}[t]{.49\textwidth}
        \centering
        \includegraphics[width=\textwidth]{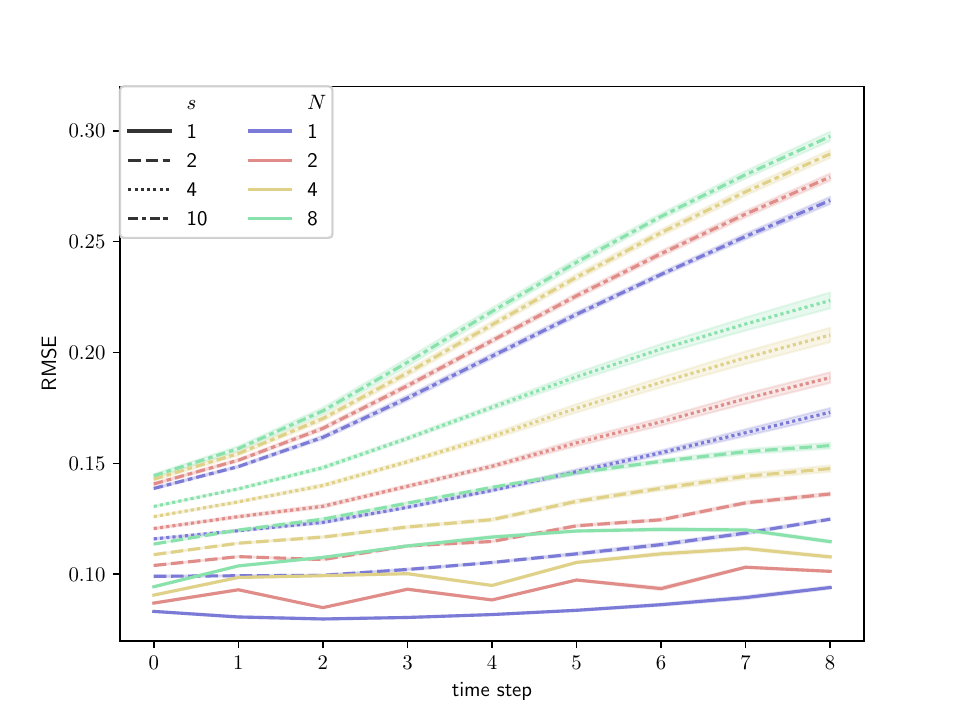}
        \caption{the easy observation $\mH^\mathrm{e}$}
        \label{fig:assimilate-with-prior-step-wise-rmse-arctan3x-soad-s}
    \end{subfigure}%
    \begin{subfigure}[t]{.49\textwidth}
        \centering
        \includegraphics[width=\textwidth]{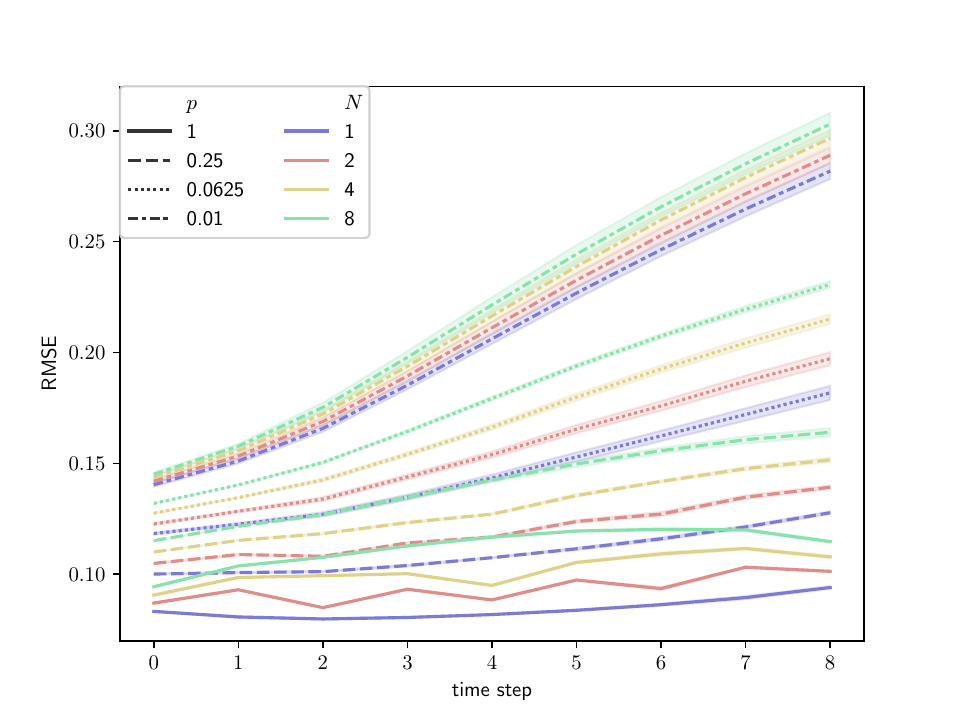}
        \caption{the easy observation $\mH^\mathrm{e}$}
        \label{fig:assimilate-with-prior-step-wise-rmse-arctan3x-soad-p}
    \end{subfigure}
    \begin{subfigure}[t]{.49\textwidth}
        \centering
        \includegraphics[width=\textwidth]{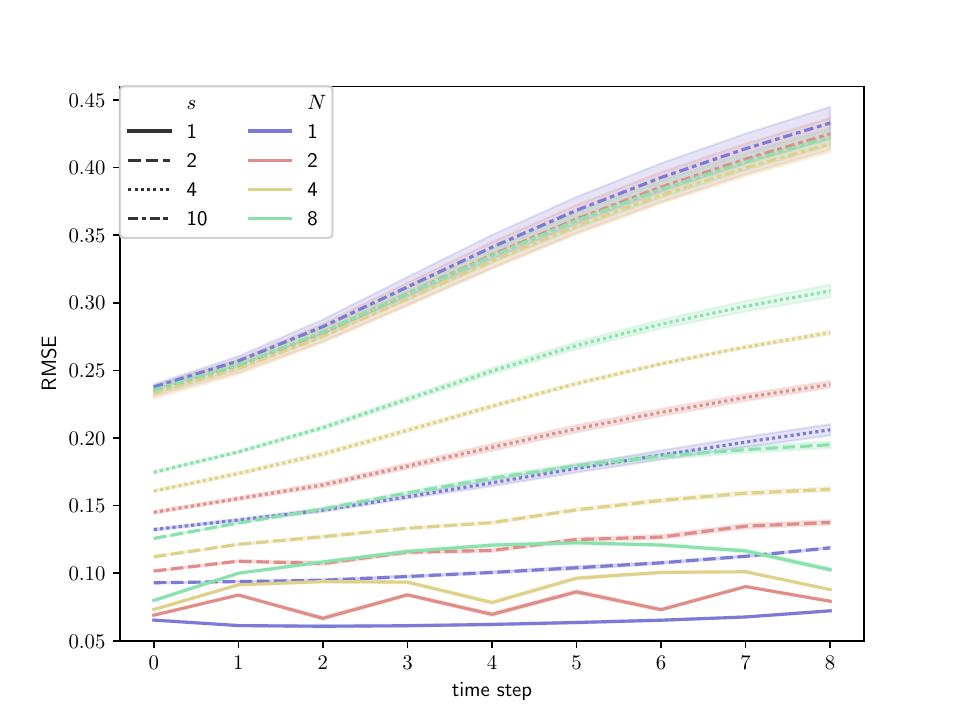}
        \caption{the hard observation $\mH^\mathrm{h}$}
        \label{fig:assimilate-with-prior-step-wise-rmse-sin3x-soad-s}
    \end{subfigure}%
    \begin{subfigure}[t]{.49\textwidth}
        \centering
        \includegraphics[width=\textwidth]{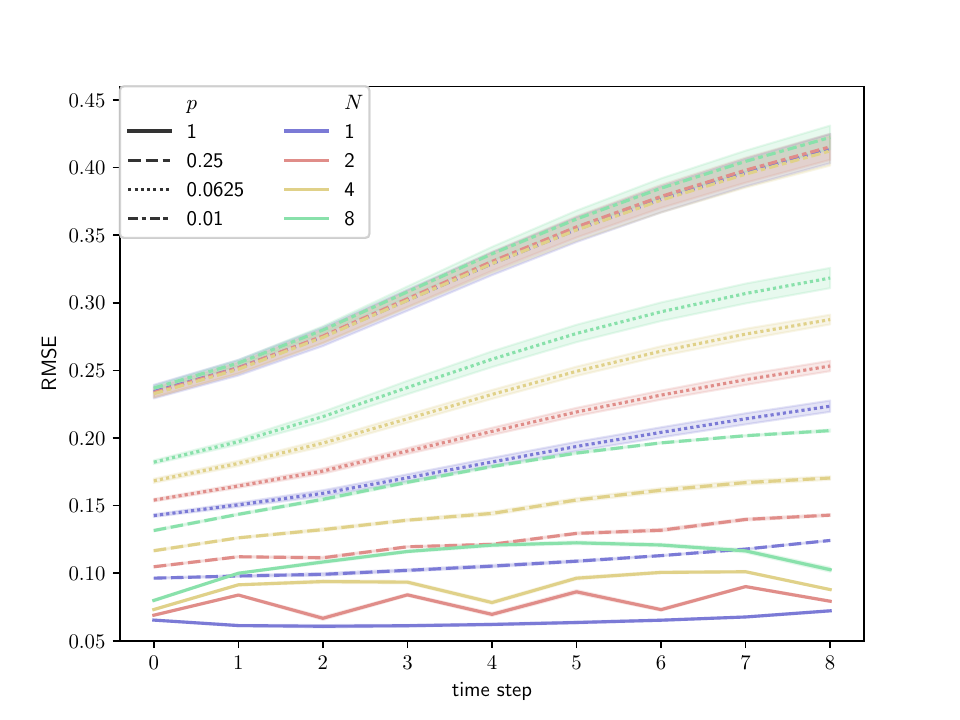}
        \caption{the hard observation $\mH^\mathrm{h}$}
        \label{fig:assimilate-with-prior-step-wise-rmse-sin3x-soad-p}
    \end{subfigure}
    \begin{subfigure}[t]{.49\textwidth}
        \centering
        \includegraphics[width=\textwidth]{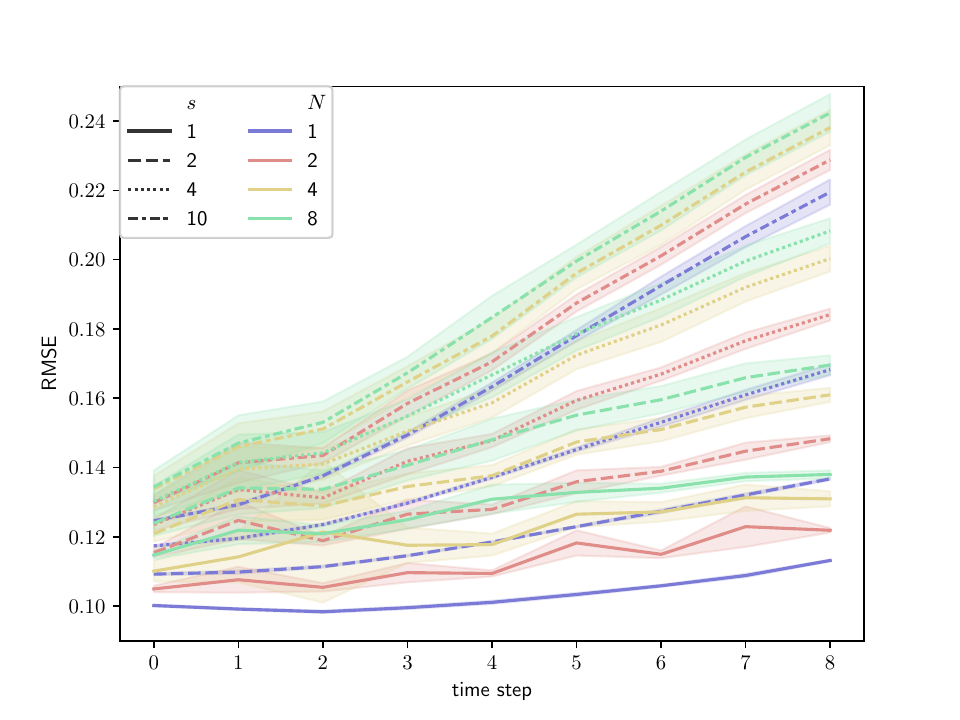}
        \caption{the vorticity-to-velocity $\mH^\mathsf{v2v}$}
        \label{fig:assimilate-with-prior-step-wise-rmse-vor2vel-soad-s}
    \end{subfigure}%
    \begin{subfigure}[t]{.49\textwidth}
        \centering
        \includegraphics[width=\textwidth]{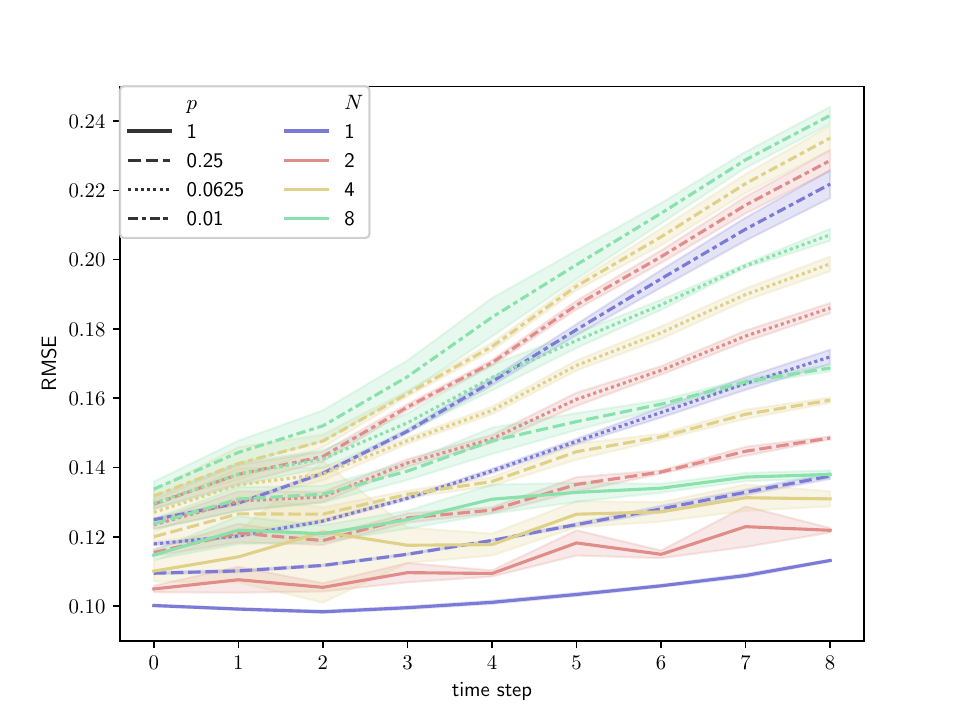}
        \caption{the vorticity-to-velocity $\mH^\mathsf{v2v}$}
        \label{fig:assimilate-with-prior-step-wise-rmse-vor2vel-soad-p}
    \end{subfigure}
    \caption{{Step-wise RMSEs of SOAD for assimilation with prior.}}
    \label{fig:assimilate-with-prior-step-wise-rmse-for-soad}
\end{figure}

Additionally, the step-wise RMSEs for our SOAD model over the assimilation window are exhibited in \cref{fig:assimilate-with-prior-step-wise-rmse-for-soad}.
{
Note that we start from a background prior as an initial estimate, which introduces much more physical information than the subsequent observational data. Although observations are available for some of the subsequent time steps, the physical information encoded in the observations may not be sufficiently informative to fully correct the trajectory as it evolves, particularly in the early stages of assimilation. This can result in increasing deviation from the ground truth at later time steps, especially if the dynamical system exhibits sensitivity to initial conditions or chaotic behavior.}
In both observation cases, The increments of RMSEs at observational ratios of 100\% and 25\% are mild, and no significant changes in RMSEs are observed when the easier $\mH^\mathrm{e}$ is replaced with the harder $\mH^\mathrm{h}$. Moreover, relatively lower assimilation errors are obtained whenever observational data are available, supporting the idea that more observations enhance the assimilation process. By comparing results with the same observational ratios ($p=(1/s)^2$), we may conclude that regular (uniform) observations are more beneficial for the assimilation process than irregular (random) ones.
{
In addition, although the step-wise RMSEs increase along with the time steps, and the uncertainty of $\mH^\mathsf{v2v}$ (\mbox{\cref{fig:assimilate-with-prior-step-wise-rmse-vor2vel-soad-s,fig:assimilate-with-prior-step-wise-rmse-vor2vel-soad-p}}) is slightly higher compared with $\mH^\mathrm{e}$ (\mbox{\cref{fig:assimilate-without-prior-step-wise-rmse-arctan3x-soad-s,fig:assimilate-without-prior-step-wise-rmse-arctan3x-soad-p}}) and $\mH^\mathrm{h}$ (\mbox{\cref{fig:assimilate-without-prior-step-wise-rmse-sin3x-soad-s,fig:assimilate-without-prior-step-wise-rmse-sin3x-soad-p}}), the subsequent experiments may offer some evidences that the assimilation process is unlikely to diverge.
}

\subsubsection{Assimilation without background prior}
To explore the assimilation performance in the absence of any background prior, we remove the modification \eqref{eq:background-prior} and conduct the assimilation process solely from observational data. See the last two rows of \cref{fig:assimilation-and-forecast}. Even if starting with a background prior, any assimilation method will gradually forget the initial state information as more observations are assimilated. Therefore, the results shown in this section are likely to reveal the long-term behaviors of the assimilation methods.

\begin{figure}
    \centering
    \begin{subfigure}[t]{.33\textwidth}
        \centering
        \includegraphics[width=\textwidth]{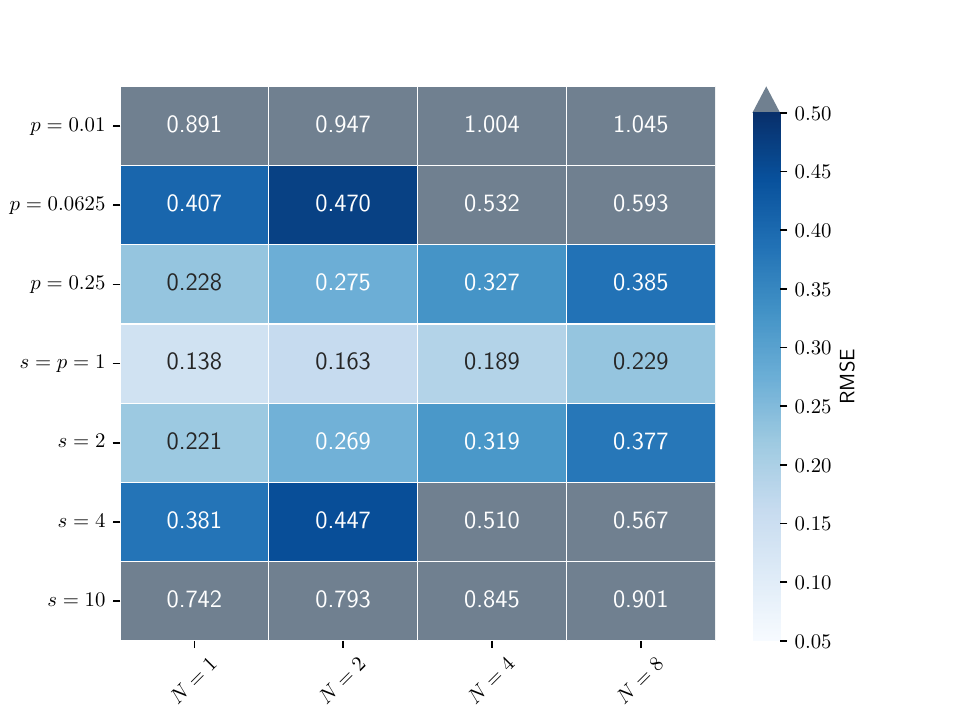}
        \caption{the easy observation $\mH^\mathrm{e}$}
        \label{fig:assimilate-without-prior-arctan3x-rmse-heatmap}
    \end{subfigure}%
    \begin{subfigure}[t]{.33\textwidth}
        \centering
        \includegraphics[width=\textwidth]{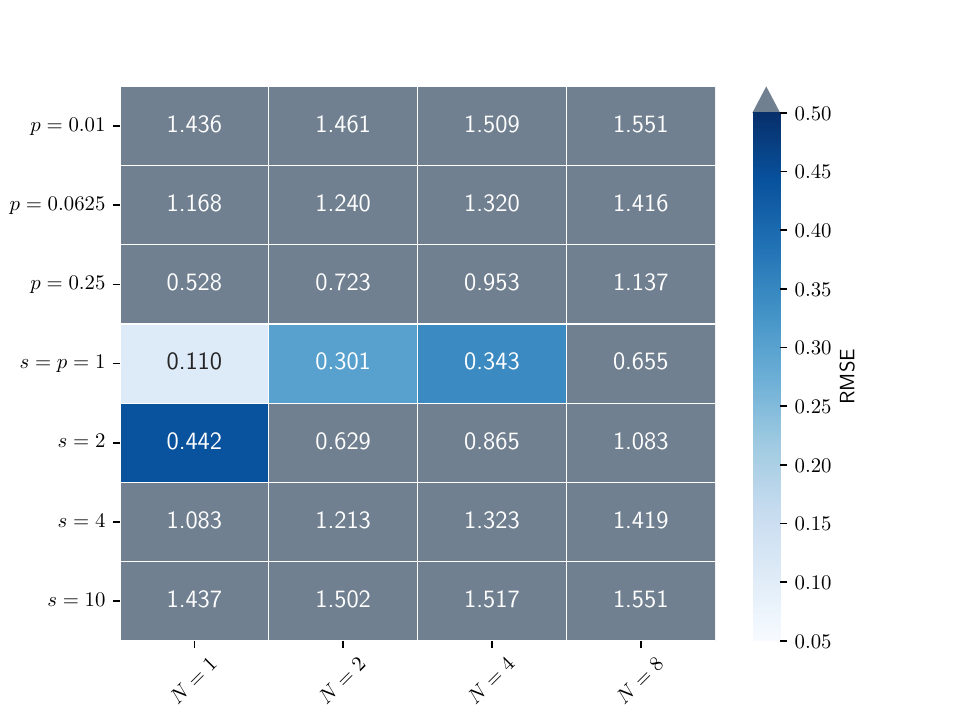}
        \caption{the hard observation $\mH^\mathrm{h}$}
        \label{fig:assimilate-without-prior-sin3x-rmse-heatmap}
    \end{subfigure}%
    \begin{subfigure}[t]{.33\textwidth}
        \centering
        \includegraphics[width=\textwidth]{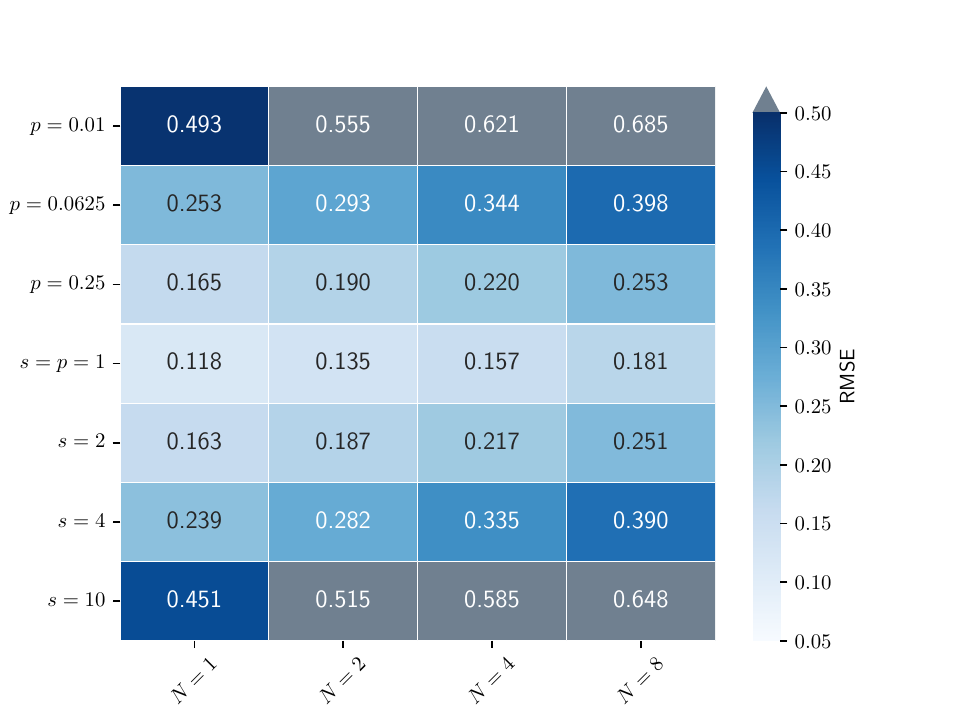}
        \caption{the vorticity-to-velocity $\mH^\mathsf{v2v}$}
        \label{fig:assimilate-without-prior-vor2vel-rmse-heatmap}
    \end{subfigure}
    \caption{Averaged RMSEs of our SOAD without background prior.}
    \label{fig:assimilate-without-prior-rmse-heatmap}
\end{figure}

All assimilation results are recorded as heatmaps of RMSEs in \cref{fig:assimilate-without-prior-rmse-heatmap}. In general, our SOAD model performs well for both {the $\mH^\mathrm{e}$ and $\mH^\mathsf{v2v}$ observational operators}, with slightly higher accuracy for the latter. The observation $\mH^\mathrm{h}$ appears to be the most challenging across all the testing observational operators. {More specifically, as also discussed in \mbox{\cref{sec:ass-with-prior}}, when observations are densely available, a condition that is uncommon in practice, the observation $\mH^\mathrm{h}$ appears to be more informative and helpful than $\mH^\mathrm{e}$ and $\mH^\mathsf{v2v}$. However, under more typical conditions with partial observations, our SOAD model achieves the best performance with $\mH^\mathsf{v2v}$, followed $\mH^\mathrm{e}$, This may be attribute to the linearity of $\mH^\mathsf{v2v}$, despite its non-local nature and greater computational complexity.}

\begin{figure}
    \centering
    \begin{subfigure}[t]{.49\textwidth}
        \centering
        \includegraphics[width=\textwidth]{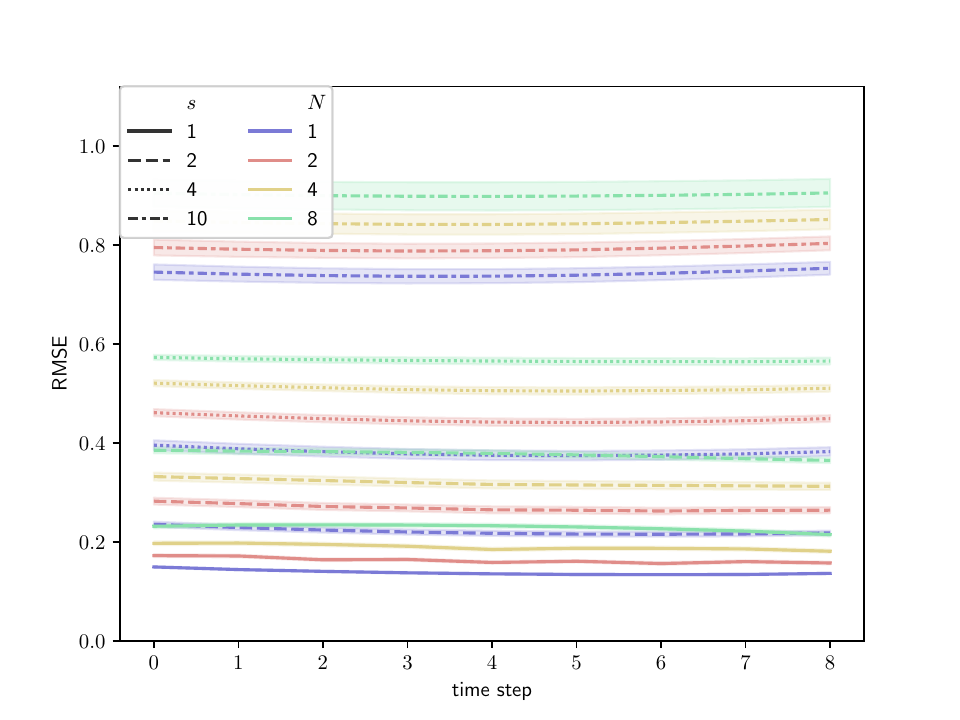}
        \caption{the easy observation $\mH^\mathrm{e}$}
        \label{fig:assimilate-without-prior-step-wise-rmse-arctan3x-soad-s}
    \end{subfigure}%
    \begin{subfigure}[t]{.49\textwidth}
        \centering
        \includegraphics[width=\textwidth]{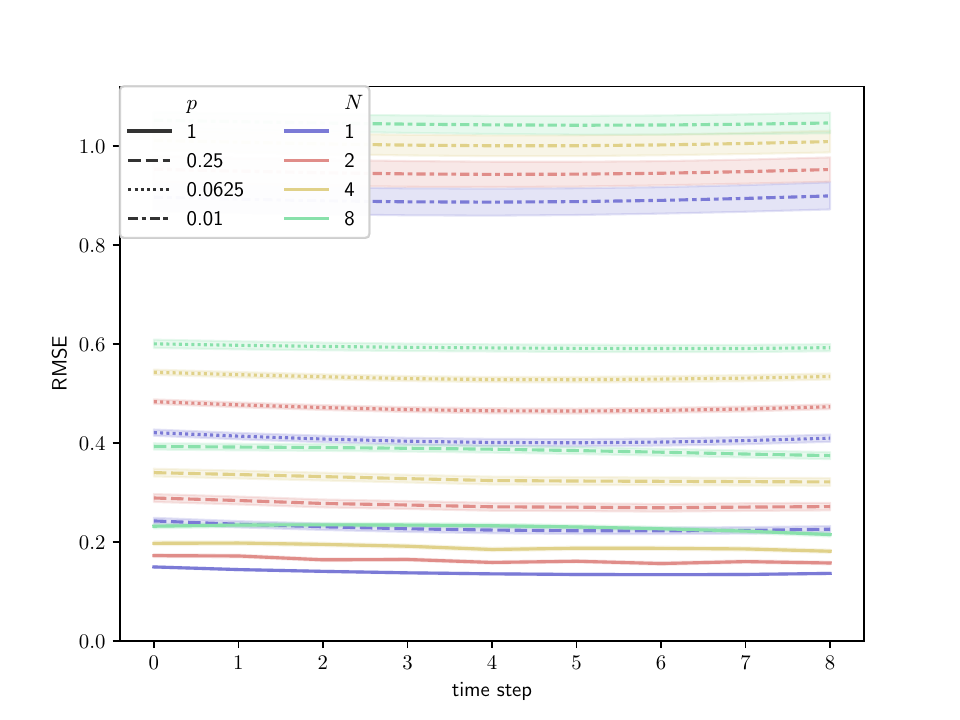}
        \caption{the easy observation $\mH^\mathrm{e}$}
        \label{fig:assimilate-without-prior-step-wise-rmse-arctan3x-soad-p}
    \end{subfigure}
    \begin{subfigure}[t]{.49\textwidth}
        \centering
        \includegraphics[width=\textwidth]{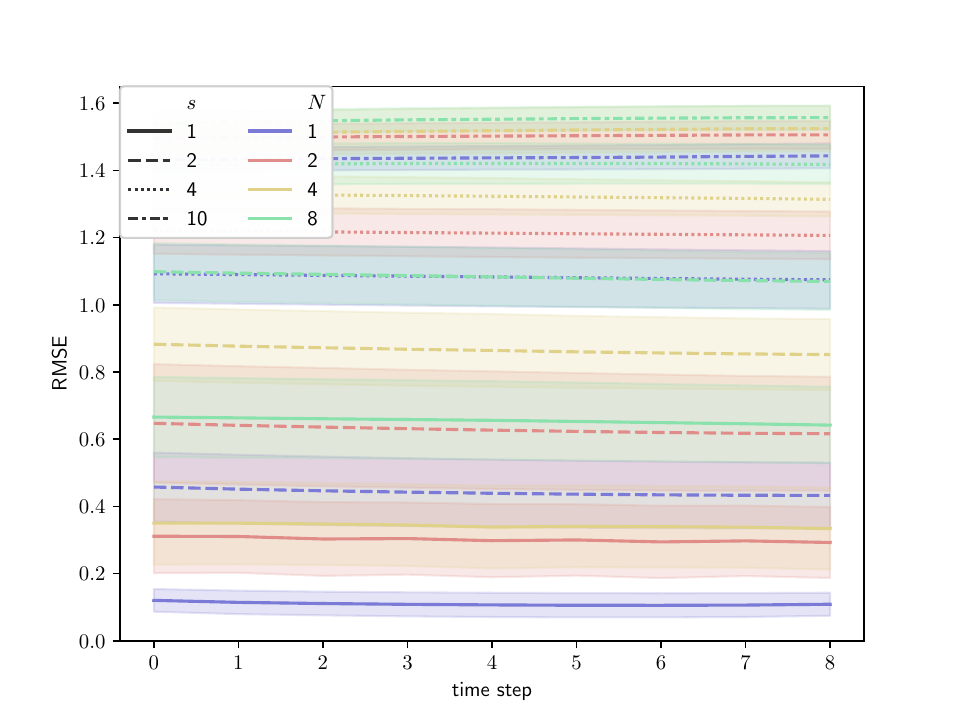}
        \caption{the hard observation $\mH^\mathrm{h}$}
        \label{fig:assimilate-without-prior-step-wise-rmse-sin3x-soad-s}
    \end{subfigure}%
    \begin{subfigure}[t]{.49\textwidth}
        \centering
        \includegraphics[width=\textwidth]{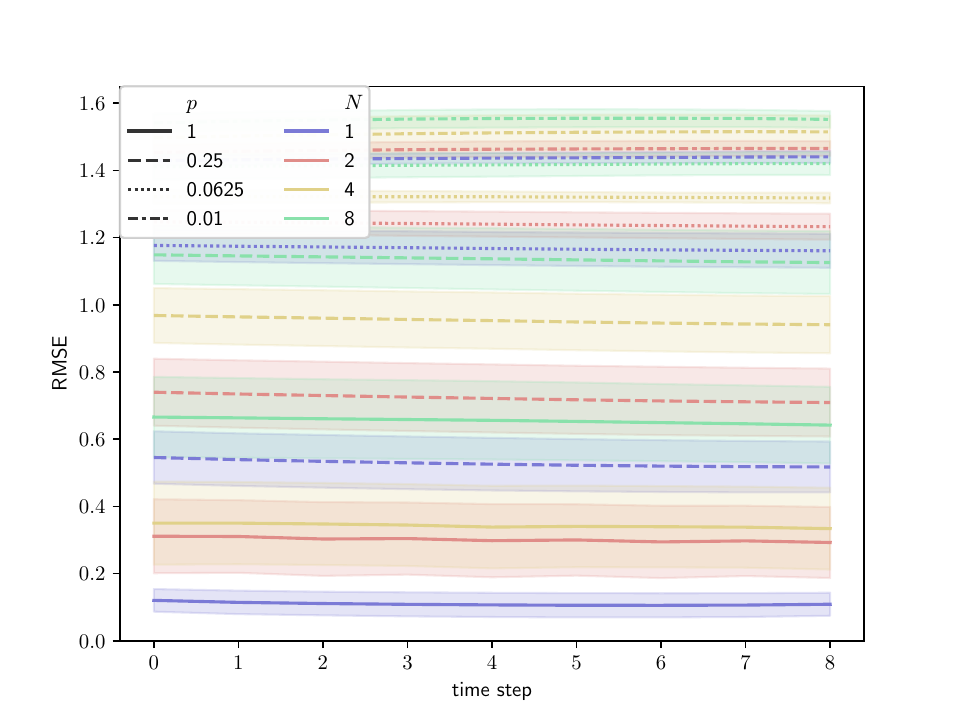}
        \caption{the hard observation $\mH^\mathrm{h}$}
        \label{fig:assimilate-without-prior-step-wise-rmse-sin3x-soad-p}
    \end{subfigure}
    \begin{subfigure}[t]{.49\textwidth}
        \centering
        \includegraphics[width=\textwidth]{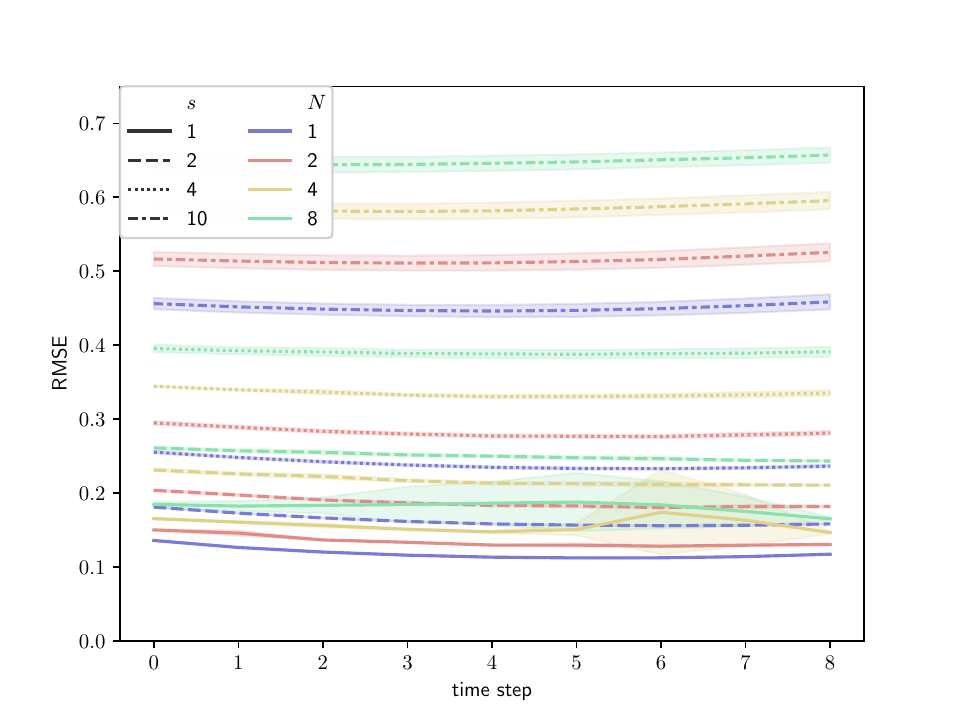}
        \caption{the vorticity-to-velocity $\mH^\mathsf{v2v}$}
        \label{fig:assimilate-without-prior-step-wise-rmse-vor2vel-soad-s}
    \end{subfigure}%
    \begin{subfigure}[t]{.49\textwidth}
        \centering
        \includegraphics[width=\textwidth]{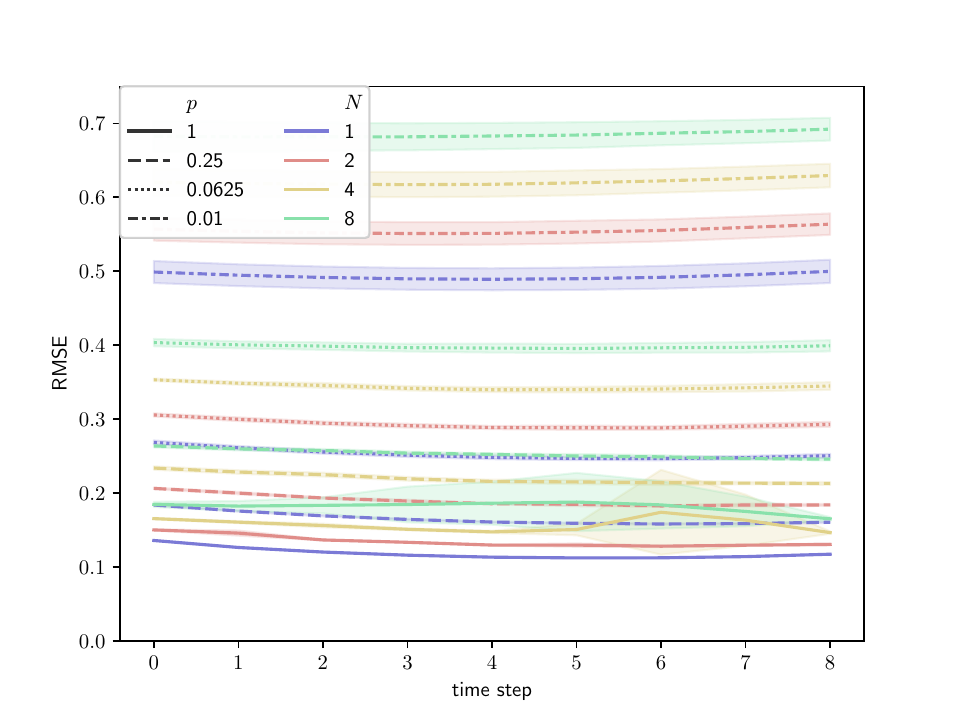}
        \caption{the vorticity-to-velocity $\mH^\mathsf{v2v}$}
        \label{fig:assimilate-without-prior-step-wise-rmse-vor2vel-soad-p}
    \end{subfigure}%
    \caption{{Step-wise RMSEs of SOAD for assimilation without prior.}}
    \label{fig:assimilate-without-prior-step-wise-rmse-soad}
\end{figure}

{
As before, we present the step-wise RMSEs of our SOAD model in \mbox{\cref{fig:assimilate-without-prior-step-wise-rmse-soad}}. Compared to the results in \mbox{\cref{fig:assimilate-with-prior-step-wise-rmse-for-soad}}, the performance degrades when assimilating without a background prior--an expected outcome given the absence of an initial estimate at the first time step. Notably, the RMSE remains relatively stable across time, with only slight increases observed at the beginning and end of the assimilation window. This pattern is consistent with the behavior of smoothing methods, which have limited observational support near the temporal boundaries. Furthermore, the individual RMSE profiles across different experimental settings demonstrate that the SOAD model maintains consistent accuracy over time, providing strong evidence of its long-term stability and resistance to divergence.
}

\begin{figure}
	\centering
	\includegraphics[width=.9\textwidth]{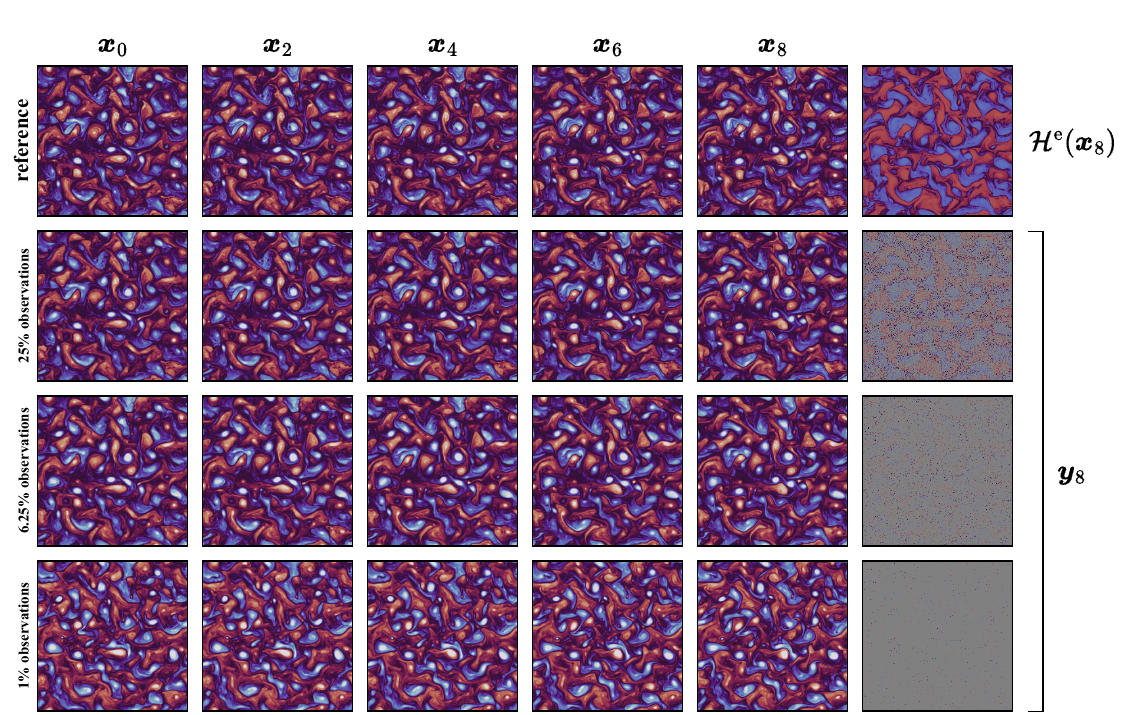}
	\caption{Visualization of the assimilated vorticities by our SOAD model with $\mH^\mathrm{e}$.}
	\label{fig:ass-without-prior-soad-arctan3x-visual}
\end{figure}

\begin{figure}
	\centering
	\includegraphics[width=.9\textwidth]{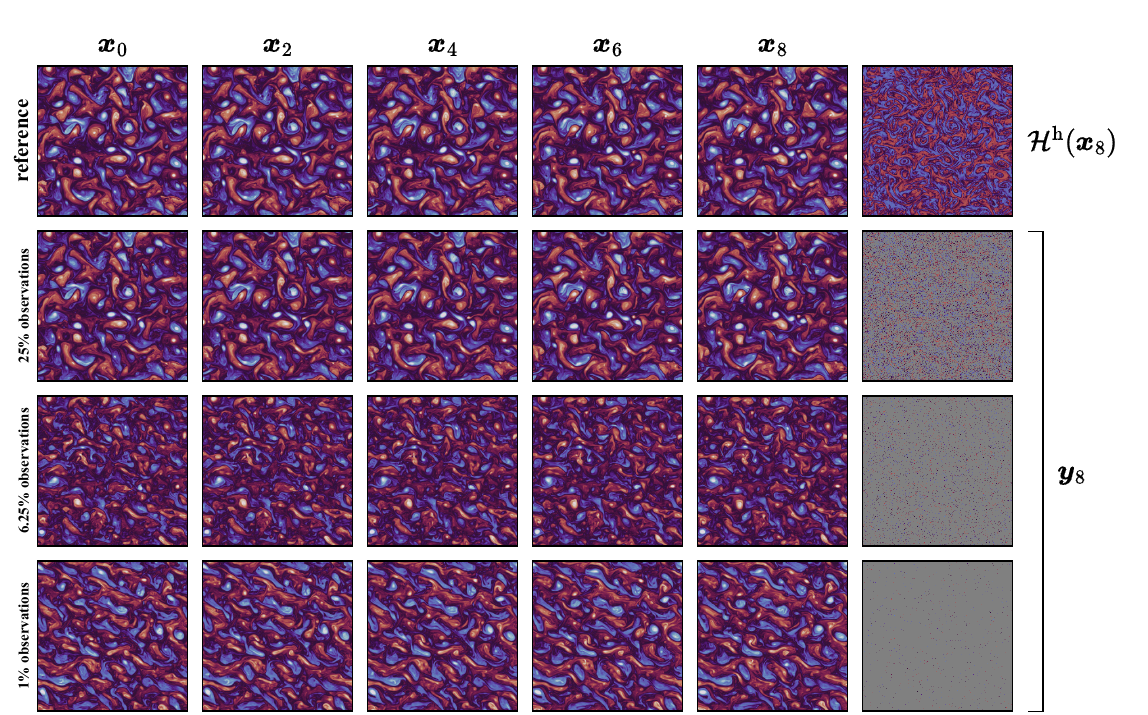}
	\caption{Visualization of the assimilated vorticities by our SOAD model with $\mH^\mathrm{h}$.}
	\label{fig:ass-without-prior-soad-sin3x-visual}
\end{figure}

\begin{figure}
	\centering
	\includegraphics[width=.9\textwidth]{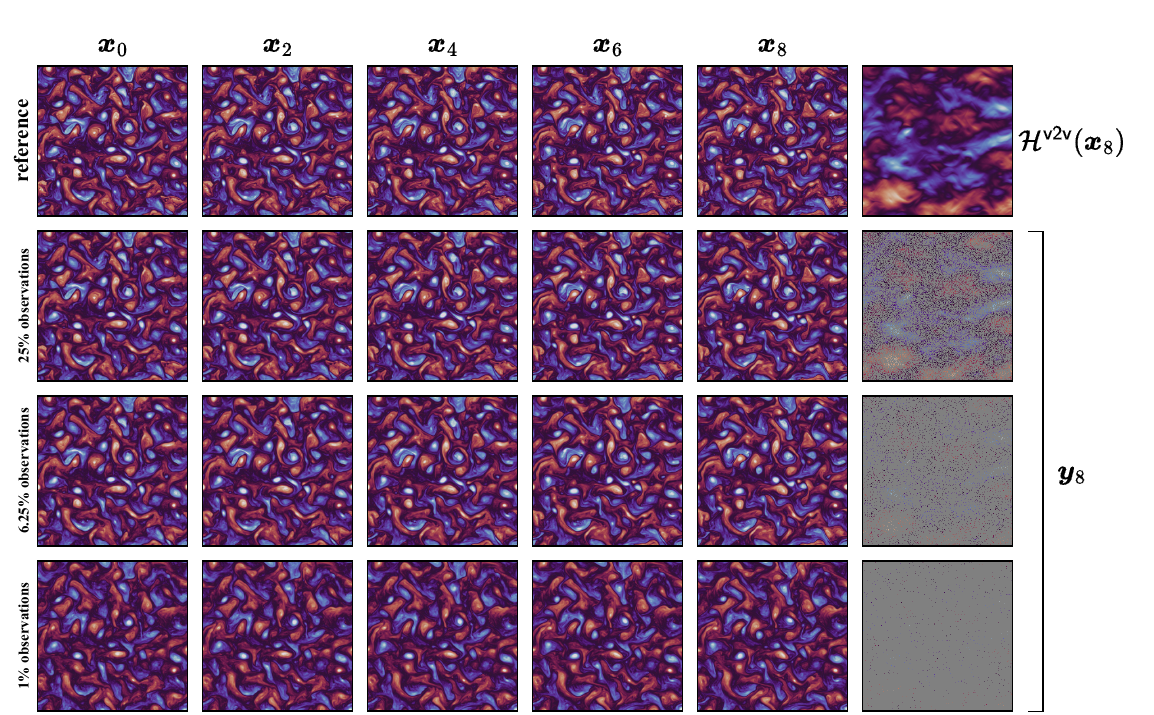}
	\caption{Visualization of the assimilated vorticities by our SOAD model with $\mH^\mathsf{v2v}$.}
	\label{fig:ass-without-prior-soad-vor2vel-visual}
\end{figure}

To visualize the assimilated states, we have plotted the assimilated vorticity with random observational ratios $p\in\{0.25,0.0625,0.01\}$ in  \cref{fig:ass-without-prior-soad-arctan3x-visual,fig:ass-without-prior-soad-sin3x-visual,fig:ass-without-prior-soad-vor2vel-visual} for $\mH^\mathrm{e}$, $\mH^\mathrm{h}$ and $\mH^\mathsf{v2v}$, respectively. For clarity, we only include the vorticity and the associated observations for the first layer, as they contain more small-scale details. For the $\mH^\mathrm{e}$ observation, our SOAD is capable of recovering most of the physical features until $p$ reaches $1\%$, in which case the assimilated states still share much similarity with the ground truth. With the harder $\mH^\mathrm{h}$, the SOAD model fails to reconstruct the real system states when $p\le0.0625=(1/4)^2$, consistent with the previous results shown in  \cref{fig:assimilate-without-prior-sin3x-rmse-heatmap}. Surprisingly, the assimilated states with rare observations do not collapse and still follow certain dynamics, suggesting that the lack of observational data, rather than incapacity of learning the dynamics, leads to the failure. Finally, when the observation is the vorticity-to-velocity mapping $\mH^\mathsf{v2v}$, the assimilated states are closer to the ground truth even when observations are available at only $1\%$ of the grids. We believe that the success may result from the linearity of $\mH^\mathsf{v2v}$. The results indicate that our SOAD approach is stable when handling highly nonlinear observations in the long term performs well on non-element-wise observations such as the vorticity-to-velocity mapping $\mH^\mathsf{v2v}$, which is a promising sign for real applications.

\subsubsection{Assimilation with multiple observations}
\begin{figure}
	\centering
	\begin{subfigure}[t]{0.49\textwidth}
		\centering
		\includegraphics[width=\textwidth]{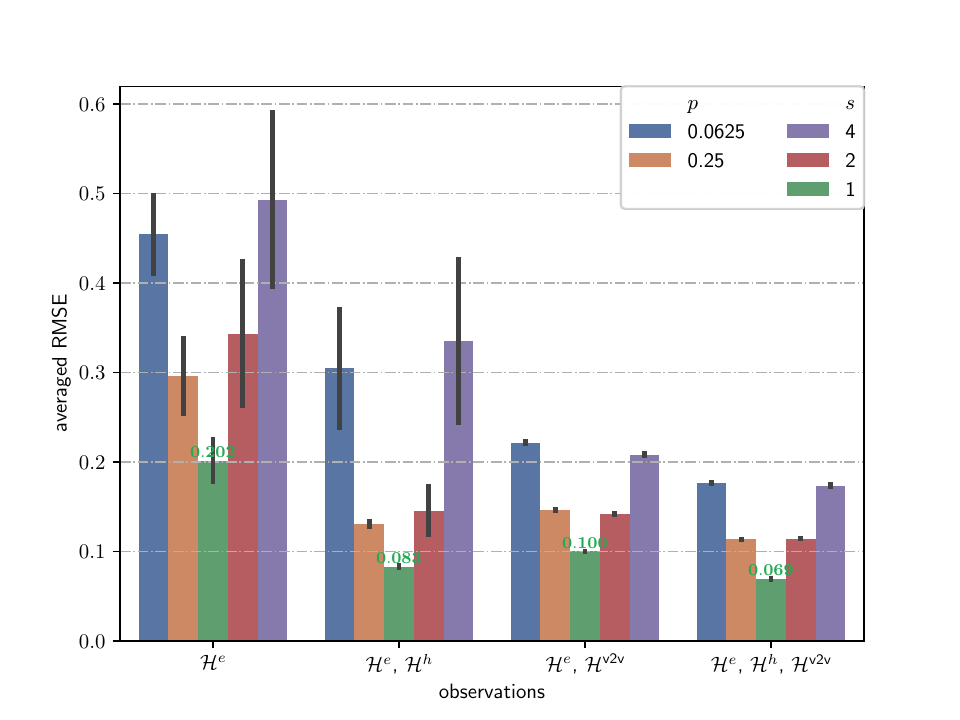}
		\caption{averaged RMSEs for collections containing $\mH^\mathrm{e}$}
	\end{subfigure}%
	\begin{subfigure}[t]{0.49\textwidth}
		\centering
		\includegraphics[width=\textwidth]{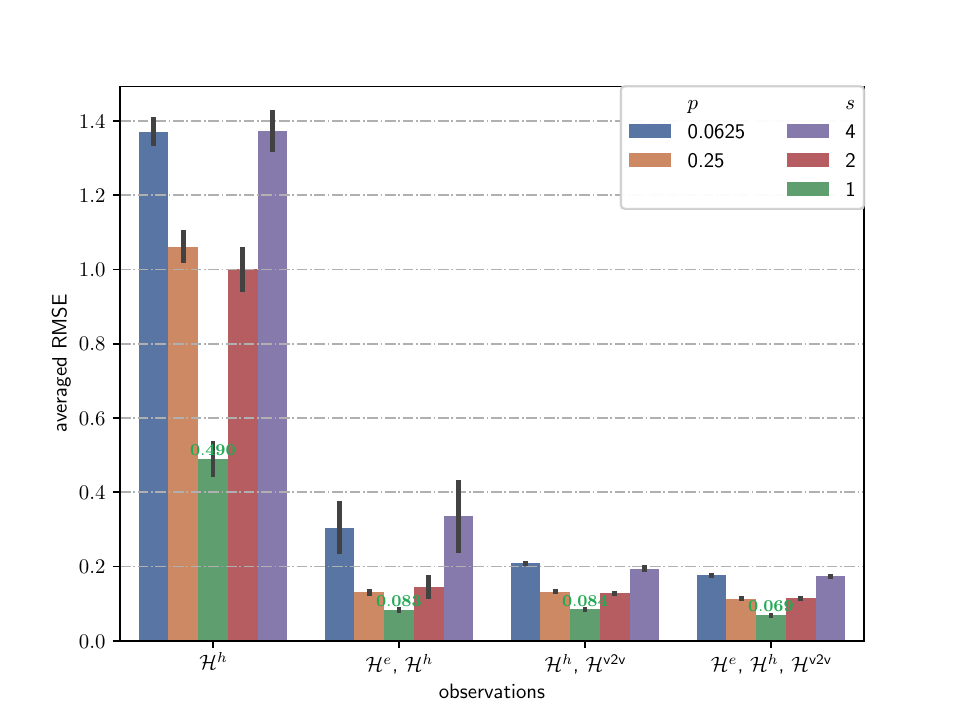}
		\caption{averaged RMSEs for collections containing $\mH^\mathrm{h}$}
	\end{subfigure}
	\begin{subfigure}[t]{0.49\textwidth}
		\centering
		\includegraphics[width=\textwidth]{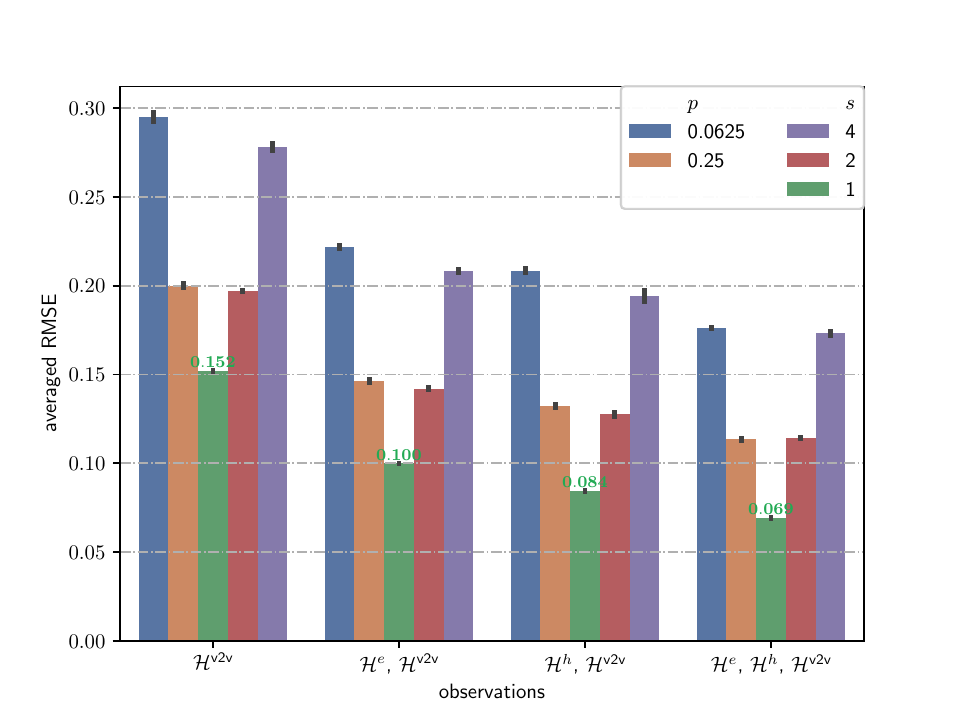}
		\caption{averaged RMSEs for collections containing $\mH^\mathsf{v2v}$}
	\end{subfigure}%
	\begin{subfigure}[t]{0.49\textwidth}
		\centering
		\includegraphics[width=.7\textwidth]{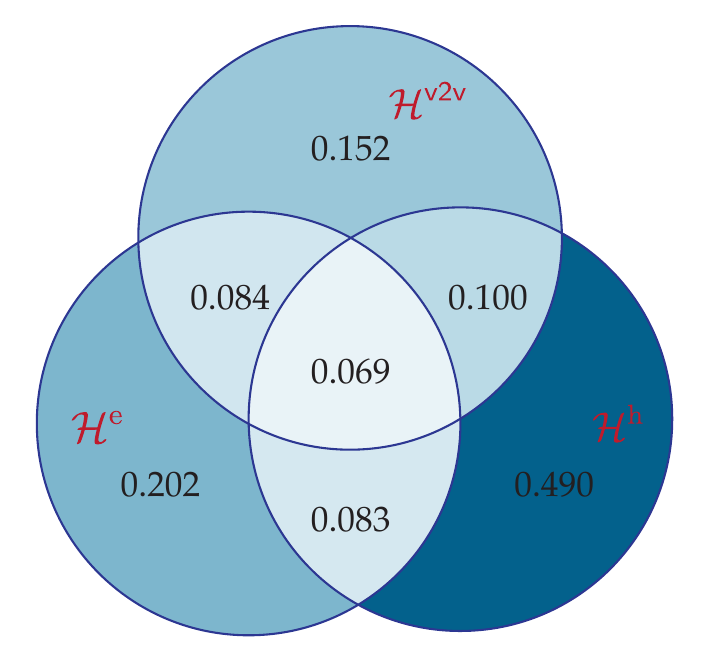}
		\caption{The best RMSEs for different combinations of observations}
		\label{fig:soad-multi-modal-rmse-Venn}
	\end{subfigure}
	\caption{RMSEs for multi-modal assimilation with our SOAD model. The error uncertainty introduced by different random seeds is marked on the top of each bar. Duplicate results may appear in multiple subplots for illustration purposes.}
	\label{fig:soad-multi-modal-rmse}
\end{figure}

In this subsection, we discuss another important feature of our SOAD approach: the ability to handle observations from different modalities. Such a capability is crucial for real applications since the observations are usually collected from various sources like satellites, weather radars, and in-situ observations. Each source may provide different physical variables in different formats. We have tested our SOAD approach with various combinations of the three observations $\mH^\mathrm{e}$, $\mH^\mathrm{h}$ and $\mH^\mathsf{v2v}$. Same as the previous subsection, we do not add any background prior for the first step in order to study the long-term behaviors. The results are displayed in \cref{fig:soad-multi-modal-rmse}.

As the collection of observations expands, the uncertainties of the assimilation errors shrink, and the averaged RMSEs decrease as well due to the additional information inferred from the observations. This phenomenon implies that our SOAD approach also follows the intuitive principle that more observations are beneficial for the assimilation process. \Cref{fig:soad-multi-modal-rmse-Venn} gathers the best performances for each observation collections when $s=p=1$, and the overlapping areas are marked with the corresponding observation combinations.

% section 5
\subsubsection{{Model robustness beyond Gaussian assumptions}}
All the previous experiments have employed Gaussian assumptions for both the background estimate (when available) and the observational noise. However, this idealization rarely holds in realistic applications. Commonly, during online deployment, the background uncertainty is either unknown or poorly characterized, and the observational noise may exhibit heavy tails or other non-Gaussian features due to sensor limitations and dynamic environmental factors. To assess the robustness of our proposed model under these more challenging but realistic conditions, we conduct additional experiments in which the Gaussian assumptions are deliberately violated. These experiments serve to evaluate the model's adaptability and reliability in the presence of distributional uncertainty.

\begin{figure}
    \centering
    \includegraphics[width=.5\textwidth]{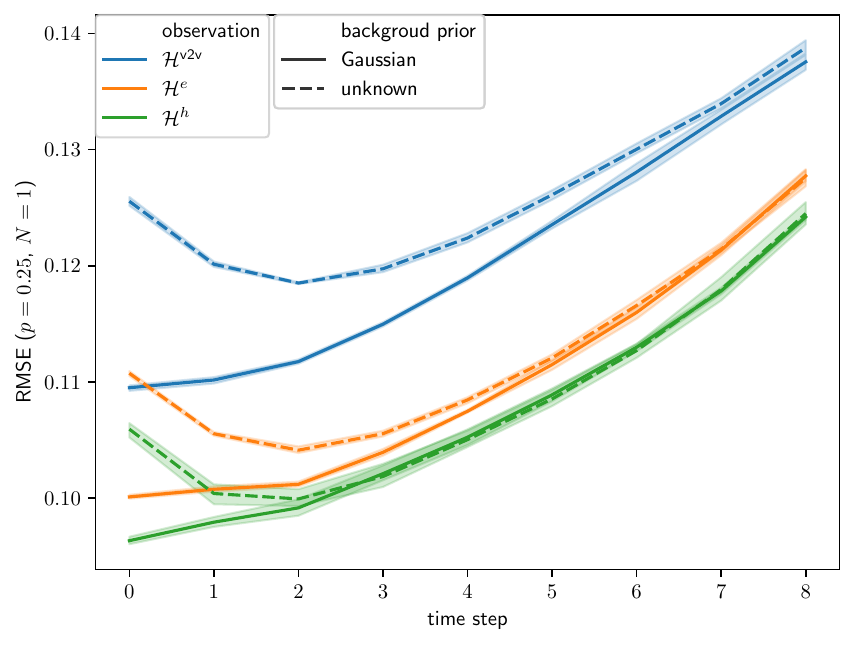}
    \caption{Step-wise RMSEs for SOAD with different data for background prior.}
    \label{fig:assimilate-background-prior-comparison}
\end{figure}

In practice, we cannot always expect the availability of the exact probability distribution for background prior. Background errors may result from many aspects, such as unresolved sub-scale features, temporal interpolations and physical parameterizations. To explore the impact of potential mismatch between our assumed background prior model \eqref{eq:background-prior} and the true underlying distribution, we design an experiment under the same settings as before, but replace the background prior with a sample generated from an unknown noise process
\begin{equation}
    \bm y_0'=\mM^\mathrm{QG}(\bm x_{-1}+\bm\eta),\quad\bm\eta\sim\mN\left(\bm0,0.01^2\bm I\right)
\end{equation}
into the model instead of $\bm y_0$. Here, $\mM^\mathrm{QG}$ is the forward propagation model of the QG equations, and $\bm x_{-1}$ stands for the state at the previous time step. Comparisons between the step-wise RMSEs are displayed in  \cref{fig:assimilate-background-prior-comparison}, where we fix $N=1$ and $p=0.25$. 
Despite the discrepancy in the background prior, our SOAD model demonstrates a capacity for self-correction within a few assimilation steps, exhibiting its robustness and relative insensitivity to inaccuracies in the background prior distribution.
% Even with incorrect background prior assumptions, our SOAD model has the ability of self-correction with a few time steps of observations, which may serve as an evidence for the robustness and insensitivity of our model to the assumptions for the background prior.

\begin{figure}
    \centering
    \includegraphics[width=0.6\linewidth]{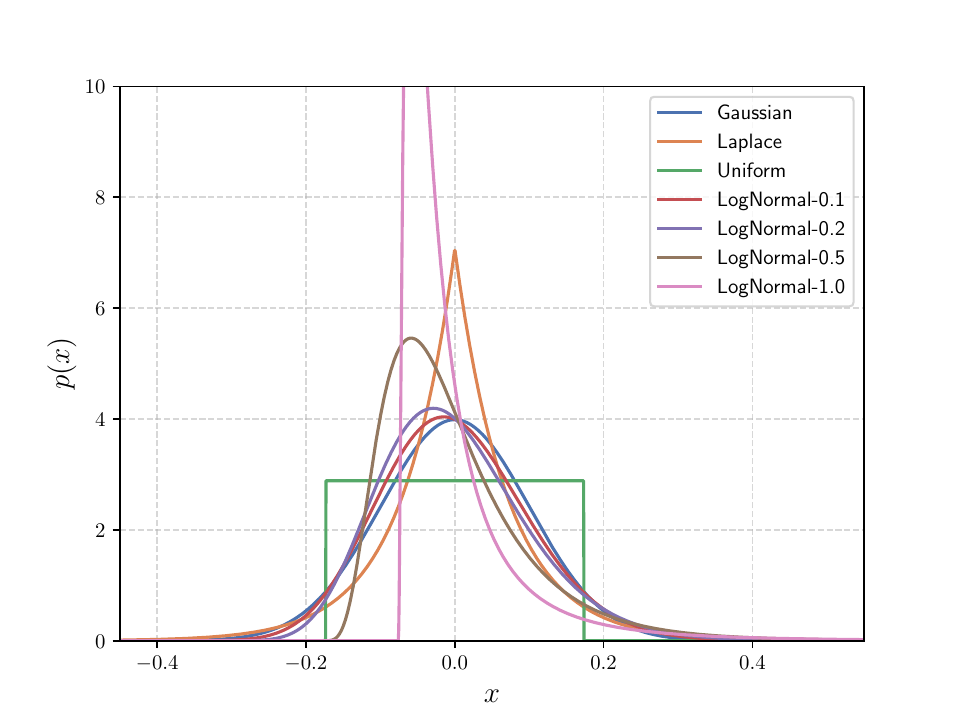}
    \caption{The probabilistic density functions for various observational noise.}
    \label{fig:various-noise-pdf}
\end{figure}

Another important source of uncertainty in data assimilation arises from the nature of observational noise, which is often assumed to be Gaussian for analytical convenience. However, in reality, observational errors can exhibit non-Gaussian characteristics due to factors such as unknown nonlinear sensor transformations, mixed error sources, or limitations in the statistical characterization of measurement systems. To evaluate the robustness of our assimilation algorithm under such conditions, we perform experiments using a non-Gaussian noise model for the observations. 
In particular, compared to \cref{eq:observational-model}, we set
\begin{equation}\label{eq:various-observational-model}
    \bm y_k=\bm S_k\circ\mH(\bm x_k)+{\bm\epsilon_k},\quad{\bm\epsilon_k}\sim\mathscr{D},
\end{equation}
where the Gaussian noise assumption is replaced by a distribution $\mathscr{D}$, drawn from three types of element-wise noise models: Laplace distribution, Uniform distribution, and LogNormal distribution, whose density functions are visualized in \cref{fig:various-noise-pdf}. Note that we have fixed appropriate parameters for each distribution so that all the distributions share the same mean and variance with our Gaussian assumption \eqref{eq:observational-model}. The detailed parameters are listed in \cref{app:various-noise} for reference.

The quantitative results for assimilation under different observational noise distributions are summarized in \cref{tab:various-observation-N1-p1,tab:various-observation-N2-p025}. Each table reports the step-wise RMSE (scaled by $10^{-2}$) under various observation operators ($\mH^\mathrm{e}$, $\mH^\mathrm{h}$, and $\mH^\mathsf{v2v}$), with and without the use of a background prior. The first table corresponds to the setting $N=1$, $p=1$, representing full spatio-temporal coverage, while the second table uses $N=2$, $p=0.25$, introducing a more difficult case. Across both settings, we observe that the performance of the SOAD model remains stable regardless of the specific noise distribution. In particular, the RMSE values under Laplace, Uniform, and moderately skewed LogNormal noise (e.g., LogNormal-0.1 to LogNormal-0.5) closely match those obtained under Gaussian noise, indicating that the model is robust to deviations from the Gaussian assumption. Only when the LogNormal noise becomes significantly skewed (LogNormal-1.0) do we see a slight degradation in performance, especially in the absence of a background prior. These results collectively suggest that SOAD is largely insensitive to the choice of noise distribution and maintains high assimilation accuracy even under non-Gaussian observational errors.

\begin{table}
    \centering
    \resizebox{\columnwidth}{!}{%
    \begin{tabular}{l||c|c|c||c|c|c}
    \toprule
        & \multicolumn{3}{c||}{assimilation w/ prior} & \multicolumn{3}{c}{assimilation w/o prior}\\
        \midrule
         dist. $\diagdown$ obs.& $\mH^\mathrm{e}$ & $\mH^\mathrm{h}$ & $\mH^\mathsf{v2v}$&$\mH^\mathrm{e}$ & $\mH^\mathrm{h}$ & $\mH^\mathsf{v2v}$\\
         \midrule
         Laplace&$8.455^{\pm0.46}$&$6.441^{\pm0.36}$&$10.31^{\pm0.49}$&$13.82^{\pm0.53}$&$14.44^{\pm1.2}$&$11.85^{\pm0.76}$\\
         Uniform&$8.449^{\pm0.46}$&$6.441^{\pm0.36}$&$10.30^{\pm0.49}$&$13.84^{\pm0.53}$&$15.91^{\pm5.6}$&$11.85^{\pm0.76}$\\
         LogNormal-0.1&$8.456^{\pm0.47}$&$6.444^{\pm0.36}$&$10.30^{\pm0.49}$&$13.82^{\pm0.53}$&$10.14^{\pm3.5}$&$11.85^{\pm0.75}$\\
         LogNormal-0.2&$8.457^{\pm0.47}$&$6.444^{\pm0.36}$&$10.31^{\pm0.49}$&$13.82^{\pm0.53}$&$13.02^{\pm5.0}$&$11.85^{\pm0.75}$\\
         LogNormal-0.5&$8.460^{\pm0.47}$&$6.445^{\pm0.36}$&$10.31^{\pm0.49}$&$13.83^{\pm0.53}$&$12.07^{\pm6.1}$&$11.85^{\pm0.75}$\\
         LogNormal-1.0&$8.529^{\pm0.48}$&$6.527^{\pm0.37}$&$10.36^{\pm0.49}$&$14.20^{\pm0.65}$&$14.07^{\pm1.6}$&$12.01^{\pm0.78}$\\
         \midrule
         \textbf{Gaussian}&$\mathbf{8.457^{\pm0.47}}$&$\mathbf{6.444^{\pm0.36}}$&$\mathbf{10.30^{\pm0.49}}$&$\mathbf{13.82^{\pm0.53}}$&$\mathbf{12.93^{\pm3.2}}$&$\mathbf{11.85^{\pm0.75}}$\\
         \bottomrule
    \end{tabular}%
    }
    \caption{RMSE ($\times10^{-2}$) with various observational noise distributions, $N=1$, $p=1$.}
    \label{tab:various-observation-N1-p1}
\end{table}

\begin{table}
    \centering
    \resizebox{\columnwidth}{!}{%
    \begin{tabular}{l||c|c|c||c|c|c}
    \toprule
        & \multicolumn{3}{c||}{assimilation w/ prior} & \multicolumn{3}{c}{assimilation w/o prior}\\
        \midrule
         dist. $\diagdown$ obs.& $\mH^\mathsf{e}$ & $\mH^\mathsf{h}$ & $\mH^\mathsf{v2v}$&$\mH^\mathsf{e}$ & $\mH^\mathsf{h}$ & $\mH^\mathsf{v2v}$\\
         \midrule
         Laplace      &$11.96^{\pm1.1}$&$12.33^{\pm1.3}$&$13.05^{\pm1.2}$&$27.50^{\pm1.0}$&$72.34^{\pm9.3}$&$18.99^{\pm0.80}$\\
         Uniform      &$11.96^{\pm1.1}$&$12.33^{\pm1.3}$&$13.05^{\pm1.2}$&$27.51^{\pm1.0}$&$72.41^{\pm9.4}$&$18.99^{\pm0.80}$\\
         LogNormal-0.1&$11.96^{\pm1.1}$&$12.34^{\pm1.3}$&$13.05^{\pm1.2}$&$27.50^{\pm1.0}$&$71.99^{\pm9.2}$&$18.98^{\pm0.80}$\\
         LogNormal-0.2&$11.96^{\pm1.1}$&$12.34^{\pm1.3}$&$13.05^{\pm1.2}$&$27.50^{\pm1.0}$&$71.45^{\pm8.4}$&$18.98^{\pm0.81}$\\
         LogNormal-0.5&$11.96^{\pm1.1}$&$12.34^{\pm1.3}$&$13.05^{\pm1.2}$&$27.51^{\pm1.0}$&$72.38^{\pm9.0}$&$18.98^{\pm0.81}$\\
         LogNormal-1.0&$12.03^{\pm1.1}$&$12.44^{\pm1.3}$&$13.09^{\pm1.2}$&$27.67^{\pm1.0}$&$72.93^{\pm7.4}$&$19.12^{\pm0.84}$\\
         \midrule
         \textbf{Gaussian}&$\mathbf{11.96^{\pm1.2}}$&$\mathbf{12.34^{\pm1.3}}$&$\mathbf{13.05^{\pm1.2}}$&$\mathbf{27.49^{\pm0.53}}$&$\mathbf{72.14^{\pm9.2}}$&$\mathbf{18.98^{\pm0.80}}$\\
         \bottomrule
    \end{tabular}%
    }
    \caption{RMSE ($\times10^{-2}$) with various observational noise distributions, $N=2$, $p=0.25$.}
    \label{tab:various-observation-N2-p025}
\end{table}
\section{Conclusion and future work}\label{sec:conclusion}
In this study, we have proposed the State-Observation Augmented Diffusion (SOAD) model, a novel data-driven data assimilation method that operates independently of classical algorithms.
Unlike many existing approaches, SOAD does not assume linearity in the physical or observational models, yet it can still recover the posterior distribution under mild conditions, suggesting a theoretical advantage. Experiments on the two-layer quasi-geostrophic model with various settings have demonstrated its good performance and potential for real-world applications.

A key feature of SOAD is its use of augmented dynamical models. Although the original assimilation problem may be highly nonlinear, we work with a linearized equivalent by augmenting the system, allowing us to focus on linear observational models.
% One of the most significant features of our SOAD is the use of augmented dynamical models. The augmented model is essentially a linearized equivalent form associated with the original assimilation task, enabling us to focus on linear observational models even if the original problem is highly nonlinear.
By leveraging a score-based generative framework, SOAD learns the prior distribution of physical states without explicit knowledge of the physical model, embedding relevant physics implicitly. We also propose a deterministic approach to estimate the adversarial gradient and introduce a forward-diffusion corrector to stabilize the assimilation process.

Further improvements and extensions of our approach are also necessary. First, analyzing convergence rates and conditions in the presence of training errors would be beneficial. Robustness and stability are particularly important in real-world applications, and broader testing on both idealized and real-world models is needed to fully assess performance.
{While our current framework assumes additive observational noise, extending it to handle non-additive or more complex noise structures would be a valuable direction for future work, as such cases may arise in certain practical scenarios.}
Besides, adapting our SOAD approach to more complex tasks beyond data assimilation could also be an exciting avenue for future research.

\section*{Data availability}
All the codes for data generation, network training and assimilation experiments are available at \url{https://github.com/zylipku/SOAD}.
\section*{Acknowledgments}
Zhuoyuan Li and Pingwen Zhang are supported in part by the National Natural Science Foundation of China (No. 12288101).
Bin Dong is supported in part by the New Cornerstone Investigator Program.
% We would like to express our gratitude to ChatGPT for polishing the authors' written text for spelling and grammar.

\section*{Declaration of generative AI and AI-assisted technologies in the writing process}
During the preparation of this work the authors used ChatGPT in order to polish the written text for spelling and grammar. After using this service, the authors reviewed and edited the content as needed and take full responsibility for the content of the publication.

\appendix
\section{Configurations of the QG model}\label{app:qg-config}
We list the parameters (as defaults in \texttt{pyqg-jax}\footnote{\url{https://pyqg-jax.readthedocs.io/en/latest/reference.models.qg_model.html}}) we have used to generate the QG dataset.
\begin{itemize}
    \item Domain size: $10^6\times10^6$;
    \item Linear drag in lower layer: $r_{ek}=5.767\times10^{-7}$;
    % \item \deleted[id=rev2]{Amplitude of the spectral spherical filter with parameter $23.6$ to obtain the small-scale dissipation;}
    \item Gravity: $g=9.81$;
    \item Gradient of Coriolis parameter: $\beta=1.5\times10^{-11}$;
    \item Deformation radius: $r_d=1.5\times10^{4}$;
    \item Layer thickness: $H_1=500$, $H_2=2000$;
    \item Upper/Lower layer flow: $U_1=0.025$, $U_2=0${;}
    \item {$F_1=r_d^{-2}/(1+\delta)$ and $F_2=\delta F_1$, where the layer thickness ratio $\delta=H_1/H_2=0.25$.}
\end{itemize}
{To obtain the small-scale dissipation ``ssd'' mentioned in \mbox{\cref{eq:qg-model}}, a highly-selective exponential filter}
\[
{
E_f=\begin{cases}
    \exp\left[-23.6(\kappa^\star-\kappa_c)^4\right],&\kappa\ge\kappa_c\\
    1,&\kappa<\kappa_c
\end{cases}
}
\]
{
is employed as a multipler on the spectral domain. The term $\kappa^\star$ stands for the non-dimensional wavenumber, and the cutoff $\kappa_c$ is set as 65\% of the Nyquist scale $\kappa_{ny}^\star=\pi$.
}
Readers may refer to \texttt{pyqg}\footnote{\url{https://pyqg.readthedocs.io/en/latest/equations/notation_twolayer_model.html}} for the detailed numerical scheme.
\section{Network architecture}\label{sec:network-arch}
\begin{figure}
    \centering
    \includegraphics[width=\textwidth]{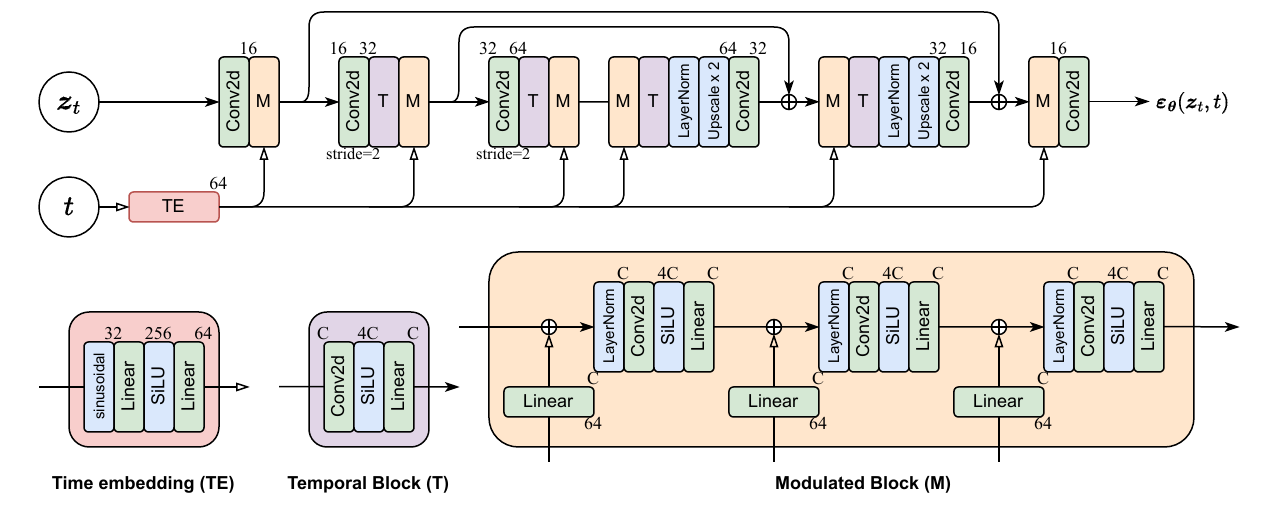}
    \caption{The U-Net architecture for the denoising network $\bm\varepsilon_\theta(\bm z_t,t)$.}
    \label{fig:unet-arch}
\end{figure}

The denoising network $\bm\varepsilon_\theta(\bm z_t,t)$ is implemented using a U-Net architecture proposed in \cite{Rozet2023sda-2lqg}, and we reproduce it in \cref{fig:unet-arch} for completeness, where we have attached the channel numbers for each module.

\section{Various observational noise distributions}\label{app:various-noise}
Let $p(x)$ denote the probabilistic density function.
To ensure all the observational noise distributions share the same mean and variance of our Gaussian assumptions $\mN\left(0,(\sigma^\mathrm{o})^2\right)$ defined in \cref{eq:observational-model}, we employ the following settings.
\begin{itemize}
    \item Laplace:
    \[p_\mathrm{Laplace}(x)=\frac1{2b}\exp\left(-\frac{|x-\mu|}{b}\right),\]
    where $\mu=0$, $b=\sigma^\mathrm{o}/\sqrt2$;
    \item Uniform:
    \[p_\mathrm{Uniform}(x)=
    \begin{cases}
        \frac1{2d},&|x|<d,\\
        0,&|x|\ge d,
    \end{cases}
    \]
    where $d=\sqrt3\sigma^\mathrm{o}$;
    \item LogNormal: Let $Z\sim\mN(\mu,s^2)$, then the distribution of $\exp Z$ is defined to follow the LogNormal distribution with parameter $\mu$ and $s$. To make the random variable unbiased, we introduce an additional shift term $c$, which means we use $(\exp Z+c)$ to generate our observational noise. For any fixed $s$, we define
    \[c=-\sigma^\mathrm{o}\left(\exp(s^2)-1\right)^{-1/2},\quad\mu=-\frac{s^2}{2}+\log(-c).\]
    The value of $s$ is appended to the distribution name in the main text.
\end{itemize}

\bibliographystyle{elsarticle-num} 
\bibliography{refs.bib}




% %% For numbered reference style

%   \textit{\LaTeX: a document preparation system},
%   Addison Wesley, Massachusetts,

% \end{thebibliography}
\end{document}